\documentclass{article} %
\usepackage[final]{colm2025_conference}

\usepackage{amsfonts}       %
\usepackage{nicefrac}       %
\usepackage{xcolor}         %
\usepackage{comment}

\usepackage{graphicx}
\usepackage{microtype}
\usepackage{hyperref}
\usepackage{url}
\usepackage{booktabs}

\usepackage{lineno}

\definecolor{darkblue}{rgb}{0, 0, 0.5}
\hypersetup{colorlinks=true, citecolor=darkblue, linkcolor=darkblue, urlcolor=darkblue}
\usepackage{xspace}
\usepackage{enumitem}
\setlist[itemize]{leftmargin=*}
\setlist[enumerate]{leftmargin=*}

\usepackage{smile}  
 
\newcommand{\Stop}{{\texttt{stop}}}
\newcommand{\Continue}{{\texttt{continue}}}
\newcommand{\ours}{\texttt{SpecDec++}\xspace}
\usepackage{mdframed}
\title{SpecDec++: Boosting Speculative Decoding via Adaptive Candidate Lengths}

\author{Kaixuan Huang \\
Princeton University  \\
\texttt{kaixuanh@princeton.edu}
\And 
Xudong Guo \\
Tsinghua University \\
\texttt{gxd20@mails.tsinghua.edu.cn}
\AND 
Mengdi Wang \\ 
Princeton University  \\
\texttt{mengdiw@princeton.edu}
}

\begin{document}

\ifcolmsubmission
\linenumbers
\fi

\maketitle

\begin{abstract}
  Speculative decoding reduces the inference latency of a target large language model via utilizing a smaller and faster draft model. Its performance depends on a hyperparameter $K$ --- the candidate length, i.e., the number of candidate tokens for the target model to verify in each round.
However, previous methods often use simple heuristics to choose $K$, which may result in sub-optimal performance. 
We study the choice of the candidate length $K$ and formulate it as a Markov Decision Process. We theoretically show that the optimal policy of this Markov decision process takes the form of a threshold policy, i.e., the current speculation should stop and be verified when the probability of getting a rejection exceeds a threshold value.
Motivated by this theory, we propose \ours, an enhanced version of speculative decoding that adaptively determines the candidate length on the fly. We augment the draft model with a trained acceptance prediction head to predict the conditional acceptance probability of the candidate tokens. \ours will stop the current speculation when the predicted probability that \textit{at least one token gets rejected} exceeds a threshold.
We implement \ours and apply it to the llama-2-chat 7B \& 70B model pair. 
Our adaptive method achieves a 2.04x speedup on the Alpaca dataset (7.2\% improvement over the baseline speculative decoding). On the GSM8K and HumanEval datasets, our method achieves a 2.26x speedup (9.4\% improvement) and 2.23x speedup (11.1\% improvement), respectively.
The code of this paper is available at \url{https://github.com/Kaffaljidhmah2/SpecDec_pp}.

\end{abstract}

\section{Introduction}
\label{sec:intro}

Current state-of-the-art Large Language Models (LLMs) have demonstrated extraordinary capabilities in various language tasks and have shown early signs of artificial general intelligence~\citep{achiam2023gpt, anil2023palm, team2023gemini, touvron2023llama, touvron2023llama2}. As the top-performing LLMs often have hundreds of billions of parameters and extremely long context windows, there is an increasing demand for serving such huge models efficiently. %

To decrease the inference latency, motivated by speculative execution techniques in processors, speculative decoding~\citep{chen2023accelerating, pmlr-v202-leviathan23a} incorporates a \textbf{draft model}, which is smaller and faster, as the speculator for the \textbf{target model}, which is the large language model we want to accelerate. Given the current prefix, the draft model first auto-regressively generates $K$ tokens, taking substantially less time than it would take the target model. The target model computes their log probabilities \textit{in parallel} and then sequentially determines whether each token is accepted or not. 
Following the first rejected token (if any), the algorithm discards the remaining tokens and corrects the rejected token with a fresh sample from a modified distribution. 
If all tokens are accepted, a new token is sampled from the next-token probability given by the target model and appended to the sequence of accepted tokens, and then the process moves forward. 
Such draft-verify-correct loops continue until the desired output is fully generated.

The speedup effect of speculative decoding depends on two crucial aspects: (1) how well the draft model aligns with the target model, and (2) how fast the draft model gets compared to the target model. The two aspects influence the choice of the hyperparameter $K$: the number of candidate tokens generated by the draft model in each loop. When the draft model aligns well and/or runs fast, we can choose a larger $K$, which potentially allows more tokens to be accepted in each loop. However, a larger $K$ also increases the chances of rejection so that more tokens get discarded.

\citet{pmlr-v202-leviathan23a} studied the problem of choosing the hyperparameter $K$ under the assumption that the acceptance rates of all the candidate tokens are constant. The authors showed that there exists one constant $K$ that can maximize the speedup.
However, such an assumption is unrealistic and does not approximate real-world cases well.
Whether the draft model and the target model align well depends on the hardness of predicting the next token. Intuitively, when the next token is unambiguous from the prefix, the draft model and the target model align well, which means the acceptance probability of the current candidate token is large compared to other cases.

\begin{figure}[t]
    \centering
    \includegraphics[width=0.95\textwidth]{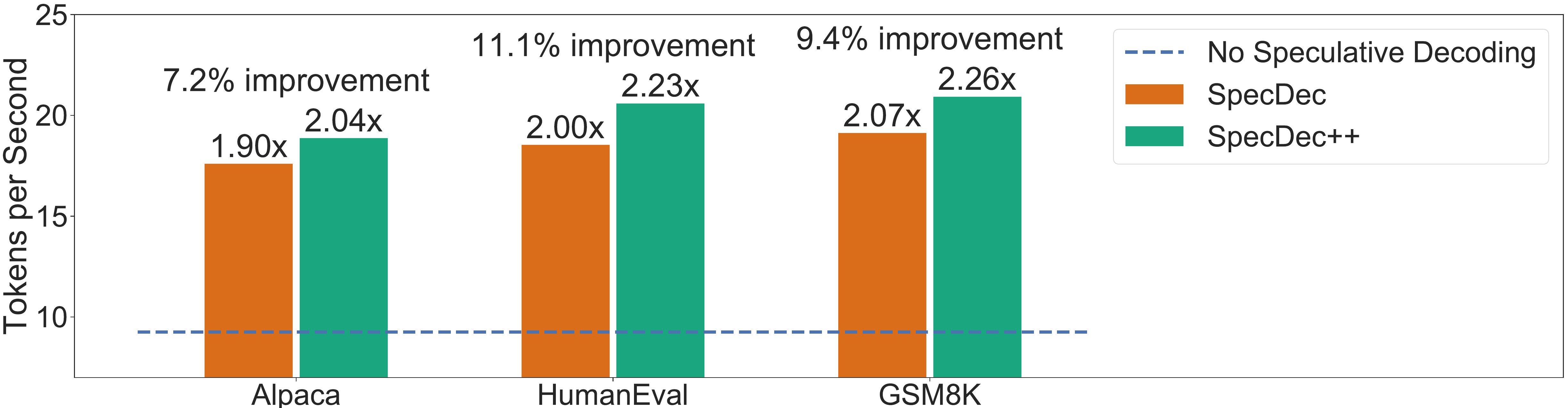} 
    \caption{The performance of \ours. Compared with the baseline speculative decoding (SpecDec) with fixed candidate lengths, by adaptively determining the candidate lengths via a trained acceptance prediction head, \ours achieves a relative \textbf{7.2\%}, \textbf{11.1\%}, and \textbf{9.4\%} improvement over the baseline methods on the Alpaca, HumanEval, and GSM8K dataset, respectively. The experiments are conducted with llama-2-chat 7B \& 70B model pair on 2 NVIDIA A100-80G GPUs.
    } 
    \label{fig:main:result}
\end{figure}

In this work, we aim to boost the performance of any speculative decoding algorithm by adaptively choosing the candidate length $K$ for each round.
We first formalize the adaptive decision-making of $K$ for speculative decoding as a Markov Decision Process (MDP). The decision to make at each timestep is whether or not to stop the current speculation round and submit the candidate tokens to the target model for verification and correction. The objective is to minimize the total inference time taken to generate a full response. Theoretically, we show that the optimal policy takes the form of a threshold policy, i.e., it is optimal to stop the speculation round whenever the probability of existing at least one rejected token in the candidates exceeds a threshold. %

Inspired by the theory, we propose \ours, an enhanced version of speculative decoding that adaptively determines the candidate length on the fly. 
First, we train an acceptance prediction head on top of the draft model to predict the acceptance probability of the candidate token. Training such an acceptance prediction head has two challenges: (1) there will be a severe class imbalance problem, e.g., most tokens generated by the draft model will have a high probability of acceptance, depending on how well the two models align; (2) the input sequence to the model contains mostly tokens from the target model and only a fraction of tokens generated by the draft model, so the training signal is sparse. 
To overcome the two challenges, we adopt a weighted Binary Cross-Entropy loss to address the class imbalance problem, and we develop a token mixing approach by randomly mixing tokens from the target model and the draft model to increase training efficiency.

At inference time, we opt to stop the current speculation round when the predicted probability of the existence of a rejected token exceeds a constant stopping threshold. The procedure is illustrated in Figure~\ref{fig:main}.
To validate the effectiveness of our proposed improvement technique, we choose the simplest implementation of speculative decoding as the baseline and augment it with \ours.
When evaluating on llama-2-chat 7B \& 70B model pair, our adaptive method achieves a 2.04x speedup compared with the 1.90x speedup of the baseline speculative decoding method on the Alpaca dataset. On the easier GSM8K and HumanEval datasets, our method boosts the baseline from 2.07x to 2.26x speedup and  from 2.00x to 2.23x speedup , respectively. 

We summarize the contributions below.
\begin{itemize}[itemsep=1pt, parsep=1pt, topsep=2pt]
    \item We formalize the dynamic choice of candidate length in speculative decoding as a Markov Decision Process (MDP) and conduct a rigorous study on the inference time. We theoretically show that when the probability that \textit{at least one token gets rejected} exceeds a threshold, the optimal action is to stop the speculation and submit it for verification.
    
    \item We propose \ours, an enhanced version of speculative decoding that adaptively determines the candidate length on the fly. We develop a weighted binary cross-entropy loss and a token mixing method to efficiently train the prediction head and use it for dynamic decision-making in the decoding process. %

    \item We validate the effectiveness of \ours with a simple baseline implementation. Our method achieves an additional 7.2\%, 9.4\%, and 11.1\% improvement over the baseline speculative decoding on the  Alpaca, HumanEval, and GSM8K datasets, respectively.
\end{itemize}

\section{Related Work}
\label{sec:related}

\textbf{Improvements on Speculative Decoding.} Since the proposal of speculative decoding, people have been improving the algorithm from different perspectives, for example, (1) making the draft model align better with the target model~\citep{zhou2024distillspec, agarwal2024policy, liu2023online}, (2) building smaller draft models or merging draft models into the target model (e.g. early-exiting)~\citep{miao2023specinfer,liu2024kangaroo,yang2023predictive,bae2023fast, zhang2024recurrent, monea2023pass, chen2023cascade}, and (3) building a heirachical system of speculative decoding~\citep{spector2023accelerating, sun2024triforce}. 

In contrast, our work focuses on the theoretical properties of the candidate length selection, and our improvement is achieved through \textbf{algorithmic improvements} that are \textit{independent of} system-level and hardware-level configurations and \textit{orthogonal to} architectural or system-level improvements.
This means \ours can be plugged into any implementation of speculative decoding as long as it adopts a form of draft models, e.g., EAGLE~\citep{li2024eagle}. Furthermore, it can be readily combined with other system and hardware improvements.

\textbf{Medusa-like Methods.} Several studies improve speculative decoding by \textit{abandoning} the auto-regressive draft model, including blockwise parallel sampling~\citep{stern2018blockwise} and the popular Medusa~\citep{cai2024medusa}. However, these approaches typically adopt a different method for verifying the candidate tokens, and the generated tokens may deviate from the target model's distribution under the general stochastic sampling setting. Therefore, we choose not to compare against this line of methods in our paper.

\textbf{Heuristic Candidate Length Selection Methods.} 
\citet{pmlr-v202-leviathan23a} make the i.i.d. assumption on the acceptance probabilities of the candidate tokens and theoretically derive the optimal choice of $K$. Besides, \citet{liu2024kangaroo} and \citet{kim2024speculative} adopt a simple heuristic that ends the speculation if the confidence of the current draft token distribution falls below a threshold. \citet{xu2023llmcad} uses the cumulative product of the confidences and extends to token trees. We include a discussion on why simple heuristics like confidence or entropy may lead to \textit{sub-optimal} performance in Appendix~\ref{appendix:discussion}.

In comparison, our work systematically studies the candidate length selection within the theoretical MDP framework and uses the cumulative product of our trained prediction head to determine the end of the speculation.  Due to space limit, please see Appendix~\ref{appendix:related} for an extended related work section.

\section{Inference Time Analysis of Speculative Decoding}
\label{sec:background}

\subsection{Background of Speculative Decoding}
To auto-regressively generate a sequence from $p(\cdot \mid x_{\text{prefix}})$ using speculative decoding, we first generate $K$ candidate tokens $(y_1, y_2, \dots, y_K)$ from $q(\cdot \mid x_{\text{prefix}})$
\[
    y_i \sim q( Y_i \mid x_{\text{prefix}}, y_1, \dots, y_{i-1}), \quad i = 1,2,\dots,K,
\]
We refer to $K$ as the candidate length, i.e., the number of candidate tokens for this round.
Next, we sequentially check if each $y_i$ is accepted or not. If there is any rejection, we replace the first rejected token with a fresh sample from the corresponding modified probability distribution and discard the subsequent tokens.  %
For completeness, the details of the speculative decoding algorithm are stated in Appendix~\ref{appendix:algo}.

The key practical consideration is that the probabilities of the candidate tokens $p(y_i \mid x_{\text{prefix}}, y_1, \dots, y_{i-1}) $ can be calculated \textit{in parallel} by the target model with no additional overhead, as the forward time is bottlenecked by the memory operations~\citep{pope2023efficiently}.
\subsection{Inference Time Decomposition of Speculative Decoding}

Our objective is to minimize the total inference time, which satisfies
\begin{equation}
        T_{\text{total}} = t_{\text{draft}}  N_{\text{draft}} + t_{\text{target}} N_{\text{target}}, \label{eq:totalcost}
\end{equation}
where  $t_{\text{draft}}$ and $t_{\text{target}}$ are the time needed for one forward pass and $N_{\text{draft}}$ and $N_{\text{target}}$ are the total number of forward passes of the draft model and the target model, respectively.
Equation~\eqref{eq:totalcost} holds under the implicit assumption that the forward passes of each of the models take constant time, which is true when we have enough computational resources to support the increased concurrency when the length of the input sequence grows~\citep{pmlr-v202-leviathan23a}. We empirically verify that Equation~\eqref{eq:totalcost} holds in our setting; see Section~\ref{sec:time}.

Let $N$ be the number of the final generated tokens. Note that $N$ is a random variable inherent to the target model and the initial prompt, independent of the draft model and the number of candidate tokens $K$ of each round we choose. Let $N_{\text{discarded}}$ be the number of total discarded tokens. By the fact that $N_{\text{draft}} + N_{\text{target}} = N + N_{\text{discarded}}$, we have the following lemma.
\begin{lemma}\label{lem:time:decompose}
The total inference time of any speculative decoding algorithm $T_{\text{total}}$ can be decomposed as
\begin{equation}
    T_{\text{total}} =  T_0  
 + t_{\text{draft}} N_{\text{discarded}} + (t_{\text{target}} - t_{\text{draft}} ) N_{\text{target}}, \label{eq:costfn}
\end{equation} 
where $T_0 = t_{\text{draft}} N $ is the oracle inference time.
\end{lemma}

To minimize the total inference time, we are required to trade-off between two objectives: minimizing the number of the discarded tokens $N_{\text{discarded}} $ and minimizing the number of forward passes of the target model $N_{\text{target}}$. The two objectives conflict with each other, as a larger candidate length $K$ will incur more discarded tokens but less number of forward passes of the target model. Equation~\eqref{eq:costfn} states that the total cost is the weighted sum of the two and the weights are given by  $t_{\text{draft}}$ and $(t_{\text{target}} - t_{\text{draft}} )$. %

\section{SpecDec++: Theory and Algorithm}
\label{sec:method}

\newcommand{\Hid}{{\boldsymbol{e}}}

\subsection{A Motivating Example: Oracle Performance of Greedy Speculative Decoding}

Let us focus on a simplified deterministic setting of speculative decoding, where we use greedy decoding for the draft model and the target model. In this setting, the draft model deterministically generates a series of greedy tokens $(Y_1,\dots, Y_K)$, and the speculative decoding algorithm reduces to sequentially checking whether $Y_i$ is also the greedy token of the target model. The first rejected token is replaced by the greedy token of the target model. If all the tokens are accepted, an additional token is generated by the target model directly.

For a given prompt $x_{\text{prompt}}$, let $(X_1,X_2,\dots,X_N)$ be the greedy tokens generated by the target model. We ask the following question:

{
    \it What is the oracle performance of the speculative decoding algorithm we can obtain by varying the number of candidate tokens, if we have the knowledge of $(X_1,X_2,\dots,X_N)$ in hindsight?
}

Let us consider the first speculation round. The draft model generates $(Y_1, Y_2, \dots)$ greedily. Let $Y_i$ be the first token such that $Y_i \ne X_i$. The optimal strategy is to stop the speculation at time $(i-1)$, so the last candidate token $Y_{i-1}$ is accepted, and $Y_i$ will be generated directly by the target model, because (1) if we stop the speculation earlier, then the shorter candidate tokens will still be accepted, but this induces at least one unnecessary forward pass of the target model; (2) if we stop the speculation later, then we waste at least one candidate token $Y_i$. By repeating the argument, we have the following.

\begin{theorem}
    In the greedy decoding setting, for a given prompt $x_{\text{prompt}}$, let $(X_1,X_2,\dots,X_N)$ be the greedy tokens generated by the target model. We define $Y_i = \argmax q(\cdot \mid x_{\text{prompt}}, X_1,X_2,\dots, X_{i-1})$ to be the greedy token of the draft model $q$ conditioned on the partial generation of the target model. Let $S$ be the set of disagreement between the draft model and the target model: $S = \{ 1 \leq i \leq N \mid Y_i \ne X_i \}$. Then, by optimally stopping at time $(i-1)$ for every $i \in S$, we obtain the oracle performance with $N_{\text{discarded}} = 0$ and $N_{\text{target}} = |S| + 1$. 
\end{theorem}

\textbf{Empirical implication.} We perform a preliminary study where we use all the prompts in the Alpaca dataset and calculate the set of disagreement $S$ for each prompt with the llama-2-chat-7B/llama-2-chat-70B model pair. The results show that the average $N_{\text{target}}/N =  0.164 \pm 0.078$ and the corresponding oracle throughput is $27.06 \pm 4.13$ tokens/second (2.92x speedup) 
in the setting of Section~\ref{sec:exp}. 
In comparison, the average throughput for the target model without speculative decoding is 9.26 tokens/second, while speculative decoding with the best fixed $K$ gives 17.58 tokens/second (1.90x speedup) (Section~\ref{sec:exp}). We can see a huge potential in adaptively tuning the candidate lengths, which motivates our subsequent study on stochastic settings and the development of \ours.

\subsection{Speculative Decoding as Markov Decision Processes} We formulate speculative decoding into the following Markov Decision Process (MDP) framework. 

\textbf{States.} We define the tuple $s = (x_{\text{prefix}}, (Y_1,\dots,Y_k))$ as the current state of the MDP. Specifically, $x_{\text{prefix}}$ is the concatenation of the prompt and the partial response containing all the accepted tokens. $(Y_1, \dots, Y_k)$ is the current candidate tokens, which are auto-regressively sampled from the draft distribution $q$:
\[
    Y_i \sim q(\cdot \mid x_{\text{prefix}}, Y_1,\dots, Y_{i-1}),\quad i=1,2,\dots.
\]
The initial state of the MDP is $(x_{\text{prompt}}, \varnothing)$.

\textbf{Actions.} Given the current state $(x_{\text{prefix}}, (Y_1,\dots,Y_k))$, the decision to make is whether or not to end the current speculation round and submit the candidate tokens to the target model for verification. We denote the current action by $a \in \{\Stop, \Continue\}$ as the choice of stopping or continuing the current speculation round. \footnote{In practice, when $Y_{k+1}$ is EOS (the special token denoting the end of sequence) or when the total length hits the maximal generation length, we manually set $a = \Stop$.}

\textbf{Transitions.} 
First, we draw a random sample $Y_{k+1} \sim q_{k+1}$ and append $Y_{k+1}$ to the current list of the candidate tokens. 

\begin{itemize}[itemsep=1pt, parsep=1pt, topsep=2pt]
    \item When $a = \Continue$, the next state $s'$ is simply $(x_{\text{prefix}}, (Y_1,\dots,Y_k, Y_{k+1}))$.
\item When $a = \Stop$, the candidate tokens $(Y_1,\dots,Y_{k+1})$ are verified via speculative decoding (Algorithm~\ref{alg:sd}). Let $n$ be the number of the accepted tokens. Let $y'$ be the replaced token when $n<k+1$ or the fresh token from the next-token distribution given by the target model when $n=k+1$. The next state $s' = ( x_\text{prefix}', \varnothing)$ with the new prefix $x_\text{prefix}' = (x_\text{prefix}, y_1,\dots,y_n,y')$ being the concatenation of the previous prefix and the newly generated tokens. 
\end{itemize}

Our next theorem provides a simple way to set the immediate cost so that the cumulative cost of the MDP matches the total inference time of Speculative Decoding.
\begin{theorem}[Immediate Cost of the MDP]
Define $c_1 = t_{\text{draft}}$ and $c_2 = (t_{\text{target}} - t_{\text{draft}} )$. If we set the immediate cost of the MDP to be
\[
    c(s, \Continue, s') = \mathbb{I}( \exists 1\leq i \leq k+1, Y_{i} \text{ is rejected}) \cdot c_1,
\]
\[
    c(s, \Stop, s') = \mathbb{I}(\exists 1\leq i \leq k+1, Y_{i} \text{ is rejected}) \cdot c_1  + c_2,
\]
then the cumulative cost of the entire trajectory equals the total inference time $T_{\text{total}}$ defined in Equation~\eqref{eq:totalcost}.
\end{theorem}

The theorem can be proved by invoking Lemma~\ref{lem:time:decompose}. For both $\Continue$ and $\Stop$, we suffer a loss $c_1$ if the current candidate token $Y_{k+1}$ is discarded, which happens if there exists any candidate token $Y_{i}$ $(1\leq i \leq k+1)$ that is rejected. If we stop at the current step, we suffer an additional cost $c_2$ corresponding to the extra inference time of the target model.

Our next theorem~\ref{thm:main} provides a sufficient condition for us to stop the current round of speculation and call the target model to verify the candidate tokens. 
\begin{theorem}\label{thm:main}
For any time-homogeneous policy $\pi$ that has an upper bound for the number of candidate tokens, at the current state $s = (x_{\text{prefix}}, (Y_1,\dots,Y_k))$, when  
\[
        \PP(\exists 1\leq i \leq {k}, Y_i \text{ is rejected} \mid x_{\text{prefix}})   \geq  \frac{c_2+\Delta}{c_1 + c_2 + \Delta},
\]
the expected total cost of $\Stop$ is smaller than the expected total cost of $\Continue$, where $\Delta = \Delta(\pi, x_{\text{prompt}}, p, q, c_1, c_2)$ is a problem-specific constant. 
\end{theorem}
We defer the proof of Theorem~\ref{thm:main} to Appendix~\ref{sec:theory}.

\subsection{\ours Algorithm}

\begin{figure}[t]
    \centering
    \includegraphics[width=0.8\textwidth]{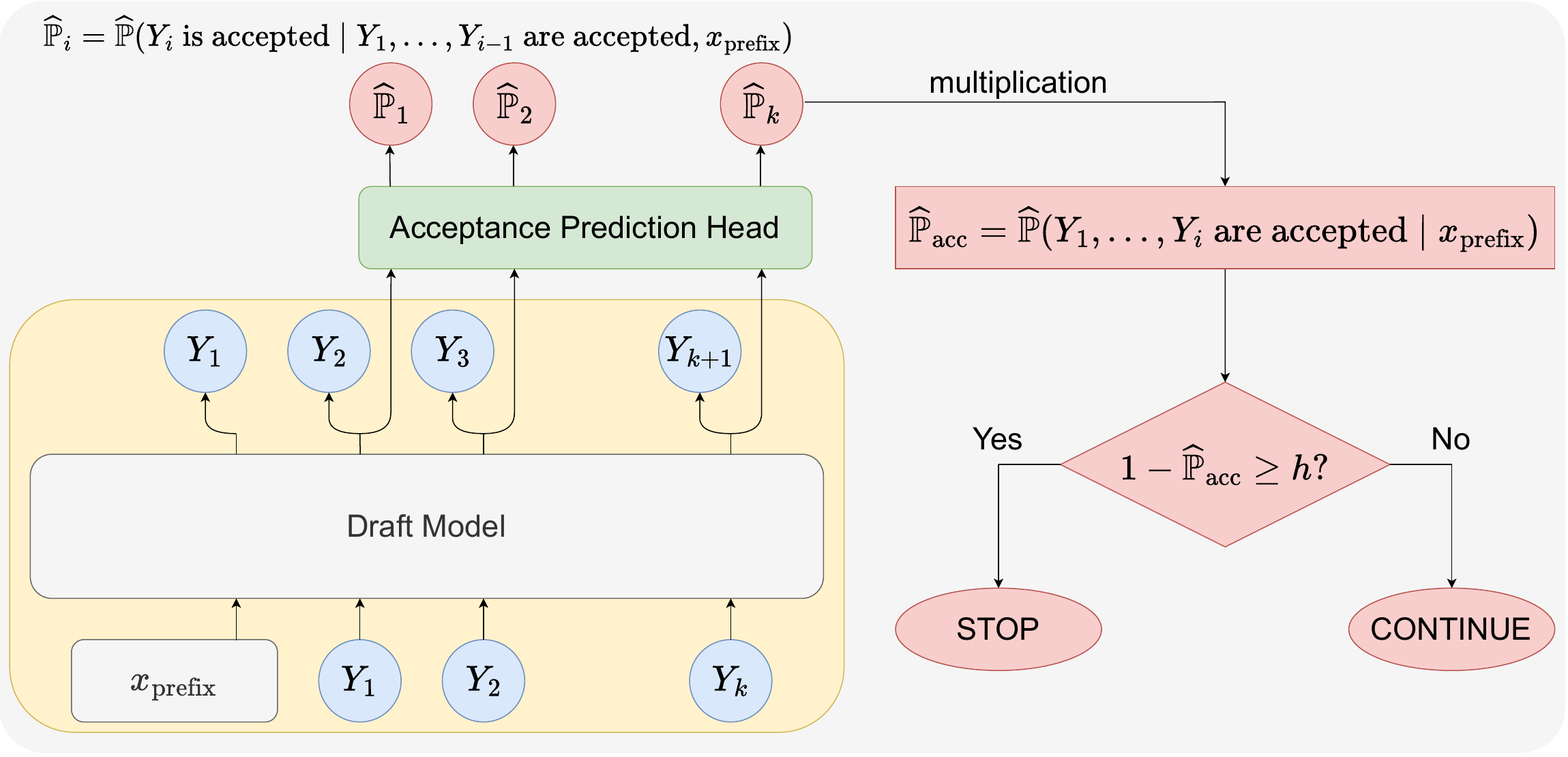} 
    \caption{\ours uses a trained \textbf{acceptance prediction head} to predict the conditional acceptance probability of the candidate tokens. When the predicted probability of the existence of at least one rejected token exceeds the \textbf{stopping threshold} $h$, the current speculation round ends and the candidate tokens go through the target model for verification and correction.
    } 
    \label{fig:main}
\end{figure}

Motivated by Theorem~\ref{thm:main}, we propose \ours, an adaptive speculative decoding algorithm that utilizes an additional prediction head to determine whether or not to stop the current speculation round. 
The additional prediction head $f_\theta$ is built on top of the draft model to predict the conditional probability 
\[
    \PP(Y_i \text{ is accepted} \mid Y_1, \dots, Y_{i-1} \text{ are accepted }, x_{\text{prefix}})  =  \min \Big(1, \frac{p(Y_i|x_{\text{prefix}}, Y_1,\dots,Y_{i-1})}{q(Y_i|x_{\text{prefix}}, Y_1,\dots,Y_{i-1})}\Big).
\] 
We opt to implement a small prediction head such that the computational overhead is negligible compared to a forward pass of the draft model. 
During inference time, we feed the input $(x_{\text{prefix}}, Y_1,\dots, Y_i)$ to the draft model and obtain the final embedding $\Hid_i$ of the last token $Y_i$. The predicted acceptance probability is given by 
\begin{equation}
    \hat{\PP}(Y_i \text{ is accepted} \mid Y_1, \dots, Y_{i-1} \text{ are accepted }, x_{\text{prefix}}) = \mathrm{sigmoid}(f_\theta(\Hid_i)). \label{eq:head}
\end{equation}
If we have such a head $f_\theta$, we propose to end the current round of speculation when the predicted probability that there exists one rejected token exceeds a predefined threshold $h$
\[
    \pi(s_k) = \Stop \Leftrightarrow  \hat{\PP} (\exists 1\leq i \leq k, \text{ such that } Y_i \text{ is rejected} \mid  x_{\text{prefix}}) > h,
\]
which can be computed by chain rule
\begin{align*}
    & \hat{\PP} (\exists 1\leq i \leq k, \text{ such that } Y_i \text{ is rejected} \mid  x_{\text{prefix}}) \\
    =& 1 - \prod_{i=1}^k   \hat{\PP}(Y_i \text{ is accepted} \mid Y_1, \dots, Y_{i-1} \text{ are accepted }, x_{\text{prefix}}).
\end{align*} 
We summarize the proposed algorithm in Algorithm~\ref{alg:sdpp} and illustrate it in Figure~\ref{fig:main}.

\begin{algorithm}[htbp]
    	\caption{\ours} %
    	\label{alg:sdpp} %
    	\begin{algorithmic} %
    	\REQUIRE draft model $q$, target model $p$, prefix $x_{\text{prefix}}$, acceptance prediction head $f_\theta$, threshold $h$. %
    \begin{mdframed}[backgroundcolor=green!20, leftmargin=5pt, rightmargin=5pt, innerleftmargin=5pt, innertopmargin=2pt, innerbottommargin=2pt, innerrightmargin=5pt]
            \STATE{\textbf{Initialize} the cumulative acceptance probability $\hat{p} = 1$}
            \FOR{$i=1$}
                \IF{$i>1$}
                    \STATE{Compute the final hidden embedding $\Hid_{i-1}$ of the token $y_{i-1}$.}
                \ENDIF 
                \STATE{Compute $q_i = q(\cdot \mid x_{\text{prefix}}, y_1, \dots, y_{i-1})$.}
                \STATE{Sample $y_i \sim q_i$.} 
                \STATE{Update $\hat{p} \gets \hat{p} \cdot \mathrm{sigmoid}(f_\theta(\Hid_{i-1}))$.}
                \IF{$1-\hat{p} > h$}
                    \STATE{\textbf{Break}}
                \ENDIF
            \ENDFOR
    \end{mdframed}
            \STATE{Let $K$ be the number of candidate tokens in the previous for-loop.} %
            \STATE{ Compute \textit{in parallel} $p_i = p(\cdot \mid x_{\text{prefix}}, y_1, \dots, y_{i-1})$ for $i=1,\dots,K+1$. }
            \STATE{ Sample $r_1,\dots,r_K$ with $r_i \sim \mathrm{Unif}[0,1]$, $i=1,\dots,K$. }
            \STATE{ Compute the number of accepted tokens $n=\min\Big (\{i-1 \mid r_i \geq p_i(y_i)/q_i(y_i)\} \cup K \Big) $.}
    	\IF{$n < K$}
                \STATE{Sample $y'$ from the modified distribution $\mathrm{Norm}[(p_{n+1} - q_{n+1})_+]$ }
            \ELSE
                \STATE{Sample $y'$ from $p_{K+1}$}
            \ENDIF
            \STATE{\textbf{Return} $x_{\text{prefix}}, y_1,\dots,y_{n}, y'$}
    	\end{algorithmic}
\end{algorithm}

\subsection{Training Dataset Construction and Learning Objective}
 \label{sec:method:train}

In this subsection, we focus on how to efficiently train an acceptance prediction head for \ours. %
For each $x_{\text{prompt}}$ in the prompt set $\mathcal{D}_{\text{prompt}}$, we first generate a target response $(X_1,\dots, X_N)$ using the target model. Next, we feed the prompt and the response into the draft model to get $q(\cdot \mid x_{\text{prompt}}, X_1,\dots, X_{i-1})$ for every $i$. 
We sample a draft candidate $Y_i$ from the distribution and calculate the conditional acceptance probability $
\PP_i = \min \Big(1, \frac{p(Y_i|x_{\text{prompt}}, X_1,\dots,X_{i-1})}{q(Y_i|x_{\text{prompt}}, X_1,\dots,X_{i-1})}\Big)$ for each token, which will be the training target for the acceptance prediction head.

\textbf{Token Mixing.} Ideally, the input to the acceptance prediction head should be $(x_{\text{prompt}}, X_1,\dots,X_{i-1}, Y_i)$. However, this naive construction is training-inefficient as only the final token $Y_i$ receives a training signal. To overcome this, we propose a token mixing strategy, borrowing the random masking idea from BERT~\citep{devlin-etal-2019-bert}: we randomly take $r\%$ tokens from  $(X_1,\dots, X_N)$ and the remaining tokens from $(Y_1,\dots, Y_N)$ to construct the response sequence, denoted by $(Z_1,\dots,Z_N)$. The losses are only computed for the tokens from $(Y_1,\dots, Y_N)$. In this way, we trade the quality of the input sequences for training efficiency, as more tokens will receive training signals per forward pass of the model.

\textbf{Weighted Binary Cross-Entropy (BCE) Loss.} 

In the typical setting of speculative decoding where the draft model and the target model align reasonably well, there will be class imbalance issues in the training dataset, where most of the training examples will have $\PP_i$ close to $1$. To accommodate the issues above, we train the prediction head using a weighted binary cross-entropy (BCE) loss, taken over the tokens $Z_i$'s stemming from $Y_i$'s. In summary, our final loss function is
\[
    \sum_{x_{\text{prompt}} \in \mathcal{D}_{\text{prompt}}} \sum_{\substack{1\leq i\leq N: \\ Z_i \text{ is taken from } Y_i}} \Big(-   w_{\text{acc}}\cdot \PP_i \log \hat{\PP}_i - w_{\text{rej}}\cdot (1-\PP_i) \log (1-\hat{\PP}_i) \Big),
\]
where $w_{\text{acc}}$ and $w_{\text{rej}}$ are the weights and $\hat{\PP}_i = \mathrm{sigmoid}(f_\theta(\Hid_i(x_{\text{prompt}}, Z_1,\dots,Z_{i-1},Y_{i})))$.

\section{Experiments}
\label{sec:exp}

\subsection{Experimental Setups}
\label{sec:exp:setup}
\textbf{Datasets and Model Pairs.} We adopt three datasets in our experiments: Alpaca~\citep{alpaca}, HumanEval~\citep{chen2021evaluating}, GSM8K~\citep{cobbe2021training}. We only use prompts of the datasets and do not use responses. In the experiments, we solely focus on llama-2-chat models~\citep{touvron2023llama2}, while in Appendix~\ref{appendix:gemma}, we also provide additional experimental results on Gemma models~\citep{team2024gemma, team2024gemma2}. We choose to use llama-2-chat 7B as the draft model and llama-2-chat 70B as the target model. To reduce memory consumption, we use the bfloat16 format for the models.

\textbf{Network Architecture, Weighted BCE Loss, and Stopping Criteria for \ours.} We build a $(D+1)$-layer ResNet with SiLU activation as the acceptance prediction head, and we sweep $D$ from $0$ (linear layer) to $4$ in the experiments.
We adopt the weighted BCE loss where set $w_\text{acc} = 1$ and choose $w_\text{rej}$ from  $\{ 1,3,6,12 \}$.
We tune the stopping threshold $h$ in $\{0.1, 0.3,0.5,0.7,0.9\}$. To ensure the robustness of \ours, we manually stop each speculation round when the number of candidate tokens exceeds $20$.

\textbf{Baseline Method.} We compare \ours with the simplest implementation of the speculative decoding algorithm where the number of the candidate tokens $K$ is fixed as a hyperparameter. We tune $K$ in $\{2,4,6,8,10, 12, 14\}$. 

\textbf{Metrics.} To measure the benefit of a speculative decoding pipeline, we divide Equation~\eqref{eq:costfn} by $N$ and get
\begin{equation}
    \text{latency} = T_{\text{total}}/N = t_{\text{draft}}   
 + t_{\text{draft}} \cdot  N_{\text{discarded}}/N + (t_{\text{target}} - t_{\text{draft}}) \cdot N_{\text{target}}/N. \label{eq:costfn2}
\end{equation} 
We report two metrics: (1) \textbf{discard rate} $N_{\text{discarded}}/N$, which is the average number of discarded tokens per one generated token, and (2) \textbf{verification rate} $N_{\text{target}}/N$, which is the average number of the forward calls of the target model per one generated token. 

Due to space limits, additional experimental setup is deferred to Appendix~\ref{appendix:exp:setup}.

\subsection{Forward Time Analysis}
\label{sec:time}
First, we verify the correctness of Equation~\eqref{eq:totalcost} and determine the forward time of the draft model $t_\text{draft}$ and the target model $t_\text{target}$ under our specific setting. We collect all the $(N_{\text{draft}}, N_{\text{target}}, T_{\text{total}})$ tuples from generations using speculative decoding (either the baseline version or \ours) and perform a linear regression to determine the coefficients. We also determine the standalone inference time when using only the draft model or the target model with linear regression. The linear regressions fit well with all $R^2 \geq 0.98$ and the results are summarized in Appendix~\ref{appendix:time}, Table~\ref{tab:time}.

\textbf{The additional cost of the acceptance prediction head is negligible}, as we find that the average $t_{\text{draft}}$ in \ours setting is \textit{smaller} than the average $t_{\text{draft}}$ in baseline SpecDec setting by $0.0004s$, which is likely caused by random noise of the environment.
Therefore, for both the baseline speculative decoding setting and \ours setting, we choose $(t_{\text{draft}}, t_{\text{target}}) = (0.0234, 0.112)$, which is the \textbf{average} between the two cases. 
For the stand-alone setting, we have $(t_{\text{draft}}, t_{\text{target}}) = (0.0207, 0.108)$, indicating that the average throughput for the target model without speculative decoding is $9.26$ tokens/second.

\subsection{Performance}

\begin{figure}[t]
    \centering
    \includegraphics[width=0.31\textwidth]{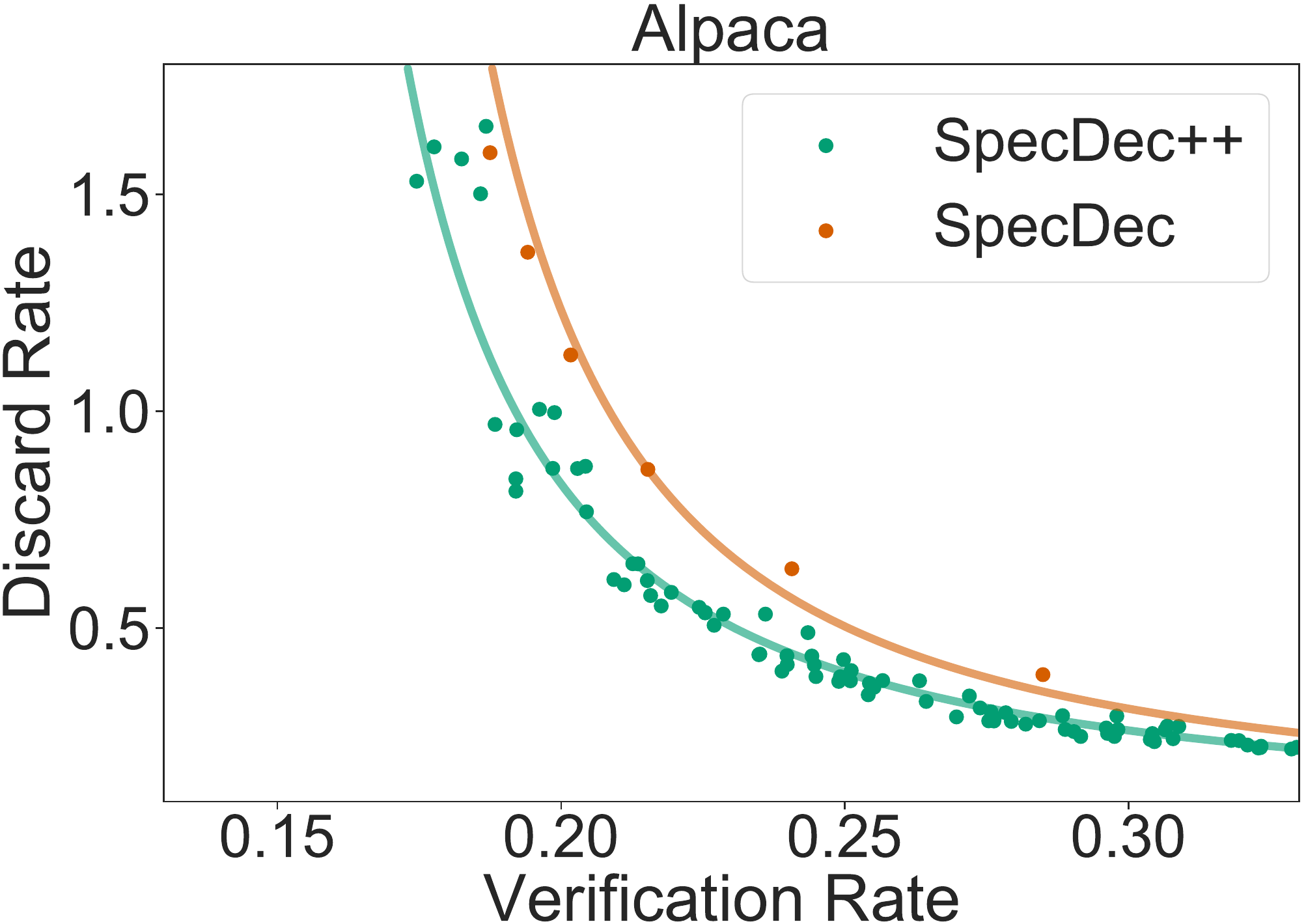}
    \includegraphics[width=0.31\textwidth]{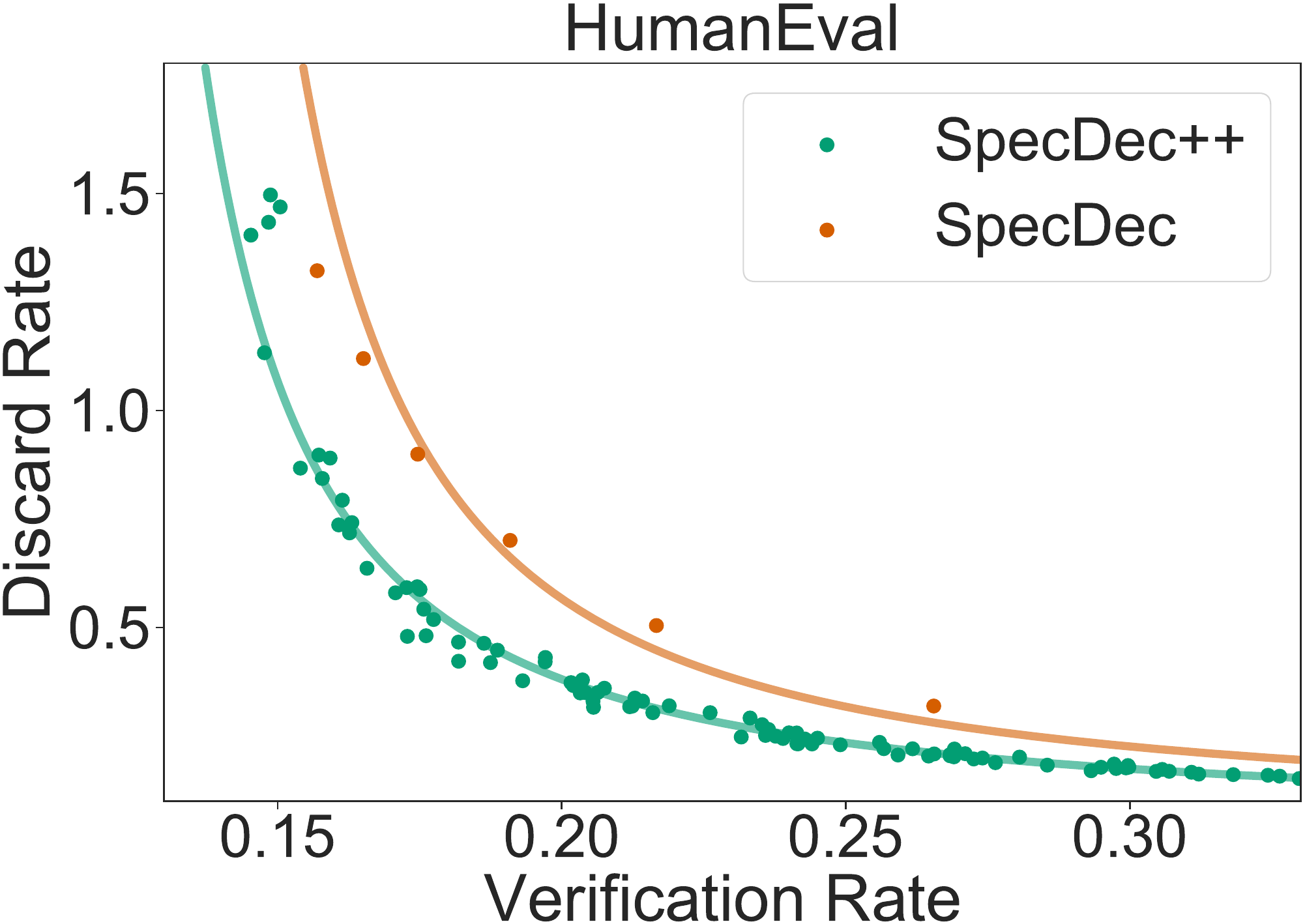}
    \includegraphics[width=0.31\textwidth]{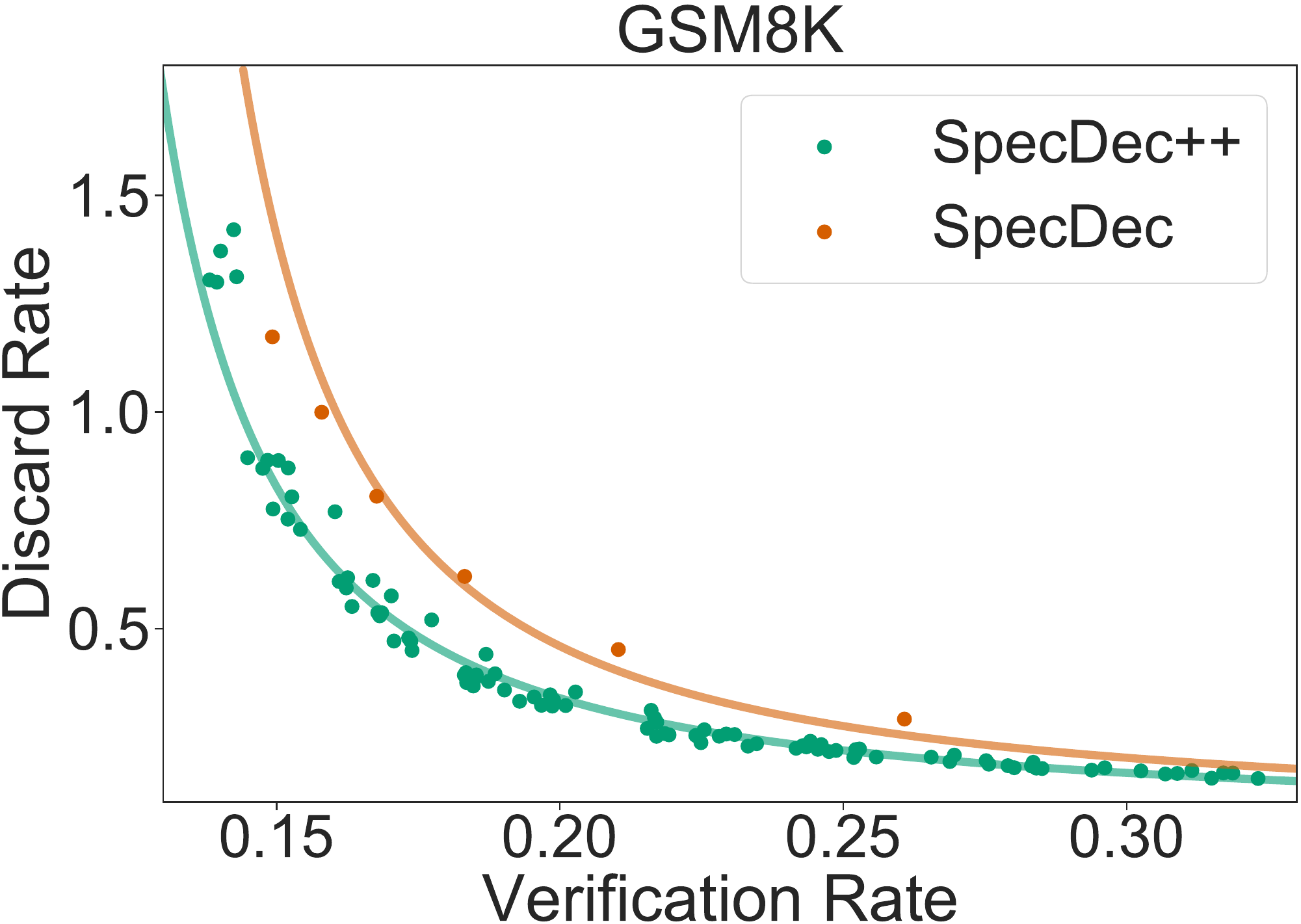}
    \caption{The average verification rates $N_{\text{target}}/N$ and the average discard rates $N_{\text{discarded}}/N$ for SpecDec with different candidate lengths and \ours with different acceptance prediction heads and stopping thresholds. \ours has better Pareto frontiers than SpecDec on both the in-distribution dataset Alpaca and the two out-of-distribution datasets HumanEval and GSM8K.
    } 
    \label{fig:pareto}
\end{figure}

We test the performances of the baseline speculative decoding with different $K$ and \ours with the different acceptance prediction heads and different thresholds $h$. We calculate the discard rates $N_\text{discarded}/N$ and the verification rates $N_\text{target}/N$ (Equation~\eqref{eq:costfn2}). The results are plotted in Figure~\ref{fig:pareto}. We see that \ours has strictly better Pareto frontiers than the baseline SpecDec on both the in-distribution test set Alpaca and the two out-of-distribution datasets HumanEval and GSM8K.
Our method with adaptive candidate lengths improves upon the baseline method of fixed candidate lengths by reducing both the discard rate and the verification rate.
The two metrics are \textbf{independent} of the actual forward times ($t_{\text{draft}}$ and $t_{\text{target}}$) and hence reusable for other hardware configurations, which indicates that \ours will still outperform the baseline under different sets of $t_{\text{draft}}$ and $t_{\text{target}}$. 
Finally, we plug in the actual values of $(t_{\text{draft}}, t_{\text{target}}) = (0.0234, 0.112)$ as in Section~\ref{sec:time}. We summarize the throughputs in Table~\ref{tab:main:result} and visualize the improvements in Figure~\ref{fig:main:result}.

\begin{table}[htbp]
    \caption{The best throughputs achieved by \ours compared to the best throughputs achieved by the speculative decoding baseline on Alpaca, HumanEval, and GSM8K datasets.}
    \label{tab:main:result}
    \centering
    \begin{tabular}{llccc}
    \hline
     Dataset   &  Alpaca  & HumanEval & GSM8K  \\ \hline
    \ours & 18.88 (tokens/s)  & 20.61  (tokens/s)   & 20.95 (tokens/s)  \\
      SpecDec (baseline)  & 17.62 (tokens/s)   &  18.55 (tokens/s)  &  19.14 (tokens/s)   \\ \hline
    \end{tabular}
\end{table}

\textbf{Discussions.} As the distribution shift of the OOD datasets will influence the accuracies and the calibrations of the acceptance prediction heads, a natural question to ask is whether the optimal performances for different datasets are achieved with different acceptance prediction heads and stopping thresholds.   Empirically, we confirm that this is indeed the case. \textit{Nevertheless}, we find that using the acceptance prediction trained with $w_{\text{rej}}=6$ and network depth $D=3$ and the stopping threshold $h=0.7$ achieves over \textbf{99.3\%} of the best tokens per second across the three datasets (2.03x for Alpaca, 2.21x for HumanEval, and 2.26x for GSM8K). Additional ablation studies on how the hyperparameters $(w_{\text{rej}}, D, h)$ influence the final tokens per second can be found in Appendix~\ref{appendix:ablation}.

\section{Conclusion and Discussion}

In this paper, we study the determination of the candidate lengths for speculative decoding. We formulate the problem as a Markov Decision Process and provide a theorem that gives a sufficient condition to stop the current speculation. Motivated by the theoretical result, we propose \ours to adaptively select the candidate length with a trained acceptance prediction head. We demonstrate significant algorithmic speedups over the naive SpecDec baselines. 
Our paper focuses on the algorithmic aspects of speculative decoding with few assumptions on the system/hardware level implementations. Therefore, our method can be seamlessly integrated with other architectural or system-level improvements.

Theoretically, the acceleration achieved via \ours depends on the learnability of the conditional acceptance probability (Eqn.~\ref{eq:head}), which, in turn, depends on the amount of correlation \textit{between} the degree of alignment between the draft distribution and the target distribution \textit{and} the existing context. While it is hard to develop intuition into when such correlation is strong for a given pair of models, our empirical results demonstrate that the correlation indeed exists and can be extracted by training the acceptance prediction head, which ultimately leads to acceleration of speculative decoding. Meanwhile, the effectivenesses of many heuristic methods~\citep{liu2024kangaroo, kim2024speculative, xu2023llmcad, mamou2024dynamic} independently support the hypothesis that such correlation can be adequate for many pairs of models.

Our final remark is on the efficacy of speculative decoding in high-throughput scenarios. The conventional wisdom suggests that speculative decoding only improves latency and may hurt throughput when batch sizes are large. However, as pointed out by \citet{sadhukhan2025magicdec}, in the prevalent \textbf{long-context} settings, KV cache loading becomes the main bottleneck of inference, and this memory bottleneck cannot be amortized by increasing batch sizes. Speculative decoding becomes an effective method to achieve speedup in this long-context scenario, even when the batch sizes are large. 
We leave for future work the adaptation of our analysis and technique to the large-batchsize, long-context settings.

\section*{Acknowledgments}

We acknowledge Tianle Cai, Kaifeng Lyu, Zhuoming Chen, and Beidi Chen for the helpful feedback and discussion.
Kaixuan Huang acknowledges the support of Google PhD Fellowship.
Mengdi Wang acknowledges support by NSF grants DMS-1953686, IIS-2107304, and ONR grant 1006977.
The research is also supported by Princeton Language and Intelligence (PLI) Compute Cluster.

\bibliography{main}

\begin{thebibliography}{54}
\providecommand{\natexlab}[1]{#1}
\providecommand{\url}[1]{\texttt{#1}}
\expandafter\ifx\csname urlstyle\endcsname\relax
  \providecommand{\doi}[1]{doi: #1}\else
  \providecommand{\doi}{doi: \begingroup \urlstyle{rm}\Url}\fi

\bibitem[Achiam et~al.(2023)Achiam, Adler, Agarwal, Ahmad, Akkaya, Aleman, Almeida, Altenschmidt, Altman, Anadkat, et~al.]{achiam2023gpt}
Josh Achiam, Steven Adler, Sandhini Agarwal, Lama Ahmad, Ilge Akkaya, Florencia~Leoni Aleman, Diogo Almeida, Janko Altenschmidt, Sam Altman, Shyamal Anadkat, et~al.
\newblock Gpt-4 technical report.
\newblock \emph{arXiv preprint arXiv:2303.08774}, 2023.

\bibitem[Agarwal et~al.(2024)Agarwal, Vieillard, Zhou, Stanczyk, Garea, Geist, and Bachem]{agarwal2024policy}
Rishabh Agarwal, Nino Vieillard, Yongchao Zhou, Piotr Stanczyk, Sabela~Ramos Garea, Matthieu Geist, and Olivier Bachem.
\newblock On-policy distillation of language models: Learning from self-generated mistakes.
\newblock In \emph{The Twelfth International Conference on Learning Representations}, 2024.

\bibitem[Anil et~al.(2023)Anil, Dai, Firat, Johnson, Lepikhin, Passos, Shakeri, Taropa, Bailey, Chen, et~al.]{anil2023palm}
Rohan Anil, Andrew~M Dai, Orhan Firat, Melvin Johnson, Dmitry Lepikhin, Alexandre Passos, Siamak Shakeri, Emanuel Taropa, Paige Bailey, Zhifeng Chen, et~al.
\newblock Palm 2 technical report.
\newblock \emph{arXiv preprint arXiv:2305.10403}, 2023.

\bibitem[Austin et~al.(2021)Austin, Johnson, Ho, Tarlow, and Van Den~Berg]{austin2021structured}
Jacob Austin, Daniel~D Johnson, Jonathan Ho, Daniel Tarlow, and Rianne Van Den~Berg.
\newblock Structured denoising diffusion models in discrete state-spaces.
\newblock \emph{Advances in Neural Information Processing Systems}, 34:\penalty0 17981--17993, 2021.

\bibitem[Bae et~al.(2023)Bae, Ko, Song, and Yun]{bae2023fast}
Sangmin Bae, Jongwoo Ko, Hwanjun Song, and Se-Young Yun.
\newblock Fast and robust early-exiting framework for autoregressive language models with synchronized parallel decoding.
\newblock In \emph{Proceedings of the 2023 Conference on Empirical Methods in Natural Language Processing}, pp.\  5910--5924, 2023.

\bibitem[Bhendawade et~al.(2024)Bhendawade, Belousova, Fu, Mason, Rastegari, and Najibi]{bhendawade2024speculative}
Nikhil Bhendawade, Irina Belousova, Qichen Fu, Henry Mason, Mohammad Rastegari, and Mahyar Najibi.
\newblock Speculative streaming: Fast llm inference without auxiliary models.
\newblock \emph{arXiv preprint arXiv:2402.11131}, 2024.

\bibitem[Cai et~al.(2024)Cai, Li, Geng, Peng, Lee, Chen, and Dao]{cai2024medusa}
Tianle Cai, Yuhong Li, Zhengyang Geng, Hongwu Peng, Jason~D. Lee, Deming Chen, and Tri Dao.
\newblock Medusa: Simple llm inference acceleration framework with multiple decoding heads.
\newblock \emph{arXiv preprint arXiv: 2401.10774}, 2024.

\bibitem[Chen et~al.(2023{\natexlab{a}})Chen, Borgeaud, Irving, Lespiau, Sifre, and Jumper]{chen2023accelerating}
Charlie Chen, Sebastian Borgeaud, Geoffrey Irving, Jean-Baptiste Lespiau, Laurent Sifre, and John Jumper.
\newblock Accelerating large language model decoding with speculative sampling.
\newblock \emph{arXiv preprint arXiv:2302.01318}, 2023{\natexlab{a}}.

\bibitem[Chen et~al.(2021)Chen, Tworek, Jun, Yuan, Pinto, Kaplan, Edwards, Burda, Joseph, Brockman, et~al.]{chen2021evaluating}
Mark Chen, Jerry Tworek, Heewoo Jun, Qiming Yuan, Henrique Ponde de~Oliveira Pinto, Jared Kaplan, Harri Edwards, Yuri Burda, Nicholas Joseph, Greg Brockman, et~al.
\newblock Evaluating large language models trained on code.
\newblock \emph{arXiv preprint arXiv:2107.03374}, 2021.

\bibitem[Chen et~al.(2024)Chen, May, Svirschevski, Huang, Ryabinin, Jia, and Chen]{chen2024sequoia}
Zhuoming Chen, Avner May, Ruslan Svirschevski, Yuhsun Huang, Max Ryabinin, Zhihao Jia, and Beidi Chen.
\newblock Sequoia: Scalable, robust, and hardware-aware speculative decoding.
\newblock \emph{arXiv preprint arXiv:2402.12374}, 2024.

\bibitem[Chen et~al.(2023{\natexlab{b}})Chen, Yang, Lin, Sun, Huang, and Chang]{chen2023cascade}
Ziyi Chen, Xiaocong Yang, Jiacheng Lin, Chenkai Sun, Jie Huang, and Kevin Chen-Chuan Chang.
\newblock Cascade speculative drafting for even faster llm inference.
\newblock \emph{arXiv preprint arXiv:2312.11462}, 2023{\natexlab{b}}.

\bibitem[Cobbe et~al.(2021)Cobbe, Kosaraju, Bavarian, Chen, Jun, Kaiser, Plappert, Tworek, Hilton, Nakano, et~al.]{cobbe2021training}
Karl Cobbe, Vineet Kosaraju, Mohammad Bavarian, Mark Chen, Heewoo Jun, Lukasz Kaiser, Matthias Plappert, Jerry Tworek, Jacob Hilton, Reiichiro Nakano, et~al.
\newblock Training verifiers to solve math word problems.
\newblock \emph{arXiv preprint arXiv:2110.14168}, 2021.

\bibitem[Devlin et~al.(2019)Devlin, Chang, Lee, and Toutanova]{devlin-etal-2019-bert}
Jacob Devlin, Ming-Wei Chang, Kenton Lee, and Kristina Toutanova.
\newblock {BERT}: Pre-training of deep bidirectional transformers for language understanding.
\newblock In Jill Burstein, Christy Doran, and Thamar Solorio (eds.), \emph{Proceedings of the 2019 Conference of the North {A}merican Chapter of the Association for Computational Linguistics: Human Language Technologies, Volume 1 (Long and Short Papers)}, pp.\  4171--4186, Minneapolis, Minnesota, June 2019. Association for Computational Linguistics.
\newblock \doi{10.18653/v1/N19-1423}.
\newblock URL \url{https://aclanthology.org/N19-1423}.

\bibitem[Du et~al.(2024)Du, Jiang, Yuanchen, Wu, Yu, Li, Li, Xu, Nie, Tu, et~al.]{du2024glide}
Cunxiao Du, Jing Jiang, Xu~Yuanchen, Jiawei Wu, Sicheng Yu, Yongqi Li, Shenggui Li, Kai Xu, Liqiang Nie, Zhaopeng Tu, et~al.
\newblock Glide with a cape: A low-hassle method to accelerate speculative decoding.
\newblock \emph{arXiv preprint arXiv:2402.02082}, 2024.

\bibitem[Fu et~al.(2024)Fu, Bailis, Stoica, and Zhang]{fu2024break}
Yichao Fu, Peter Bailis, Ion Stoica, and Hao Zhang.
\newblock Break the sequential dependency of llm inference using lookahead decoding.
\newblock \emph{arXiv preprint arXiv:2402.02057}, 2024.

\bibitem[He et~al.(2023)He, Zhong, Cai, Lee, and He]{he2023rest}
Zhenyu He, Zexuan Zhong, Tianle Cai, Jason~D Lee, and Di~He.
\newblock Rest: Retrieval-based speculative decoding.
\newblock \emph{arXiv preprint arXiv:2311.08252}, 2023.

\bibitem[Jeon et~al.(2024)Jeon, Gagrani, Goel, Park, Lee, and Lott]{jeon2024recursive}
Wonseok Jeon, Mukul Gagrani, Raghavv Goel, Junyoung Park, Mingu Lee, and Christopher Lott.
\newblock Recursive speculative decoding: Accelerating llm inference via sampling without replacement.
\newblock \emph{arXiv preprint arXiv:2402.14160}, 2024.

\bibitem[Kim et~al.(2024)Kim, Mangalam, Moon, Malik, Mahoney, Gholami, and Keutzer]{kim2024speculative}
Sehoon Kim, Karttikeya Mangalam, Suhong Moon, Jitendra Malik, Michael~W Mahoney, Amir Gholami, and Kurt Keutzer.
\newblock Speculative decoding with big little decoder.
\newblock \emph{Advances in Neural Information Processing Systems}, 36, 2024.

\bibitem[Kou et~al.(2024)Kou, Hu, He, Deng, and Zhang]{kou2024cllms}
Siqi Kou, Lanxiang Hu, Zhezhi He, Zhijie Deng, and Hao Zhang.
\newblock Cllms: Consistency large language models.
\newblock \emph{arXiv preprint arXiv:2403.00835}, 2024.

\bibitem[Kwon et~al.(2023)Kwon, Li, Zhuang, Sheng, Zheng, Yu, Gonzalez, Zhang, and Stoica]{kwon2023efficient}
Woosuk Kwon, Zhuohan Li, Siyuan Zhuang, Ying Sheng, Lianmin Zheng, Cody~Hao Yu, Joseph Gonzalez, Hao Zhang, and Ion Stoica.
\newblock Efficient memory management for large language model serving with pagedattention.
\newblock In \emph{Proceedings of the 29th Symposium on Operating Systems Principles}, pp.\  611--626, 2023.

\bibitem[Leviathan et~al.(2023)Leviathan, Kalman, and Matias]{pmlr-v202-leviathan23a}
Yaniv Leviathan, Matan Kalman, and Yossi Matias.
\newblock Fast inference from transformers via speculative decoding.
\newblock In Andreas Krause, Emma Brunskill, Kyunghyun Cho, Barbara Engelhardt, Sivan Sabato, and Jonathan Scarlett (eds.), \emph{Proceedings of the 40th International Conference on Machine Learning}, volume 202 of \emph{Proceedings of Machine Learning Research}, pp.\  19274--19286. PMLR, 23--29 Jul 2023.
\newblock URL \url{https://proceedings.mlr.press/v202/leviathan23a.html}.

\bibitem[Li et~al.(2022)Li, Thickstun, Gulrajani, Liang, and Hashimoto]{li2022diffusion}
Xiang Li, John Thickstun, Ishaan Gulrajani, Percy~S Liang, and Tatsunori~B Hashimoto.
\newblock Diffusion-lm improves controllable text generation.
\newblock \emph{Advances in Neural Information Processing Systems}, 35:\penalty0 4328--4343, 2022.

\bibitem[Li et~al.(2024{\natexlab{a}})Li, Huang, Yang, Venkitesh, Locatelli, Ye, Cai, Lewis, and Chen]{li2024snapkv}
Yuhong Li, Yingbing Huang, Bowen Yang, Bharat Venkitesh, Acyr Locatelli, Hanchen Ye, Tianle Cai, Patrick Lewis, and Deming Chen.
\newblock Snapkv: Llm knows what you are looking for before generation.
\newblock \emph{arXiv preprint arXiv:2404.14469}, 2024{\natexlab{a}}.

\bibitem[Li et~al.(2024{\natexlab{b}})Li, Wei, Zhang, and Zhang]{li2024eagle}
Yuhui Li, Fangyun Wei, Chao Zhang, and Hongyang Zhang.
\newblock Eagle: Speculative sampling requires rethinking feature uncertainty.
\newblock \emph{arXiv preprint arXiv:2401.15077}, 2024{\natexlab{b}}.

\bibitem[Liu et~al.(2024)Liu, Tang, Liu, Ni, Han, and Wang]{liu2024kangaroo}
Fangcheng Liu, Yehui Tang, Zhenhua Liu, Yunsheng Ni, Kai Han, and Yunhe Wang.
\newblock Kangaroo: Lossless self-speculative decoding via double early exiting.
\newblock \emph{arXiv preprint arXiv:2404.18911}, 2024.

\bibitem[Liu et~al.(2023)Liu, Hu, Bailis, Stoica, Deng, Cheung, and Zhang]{liu2023online}
Xiaoxuan Liu, Lanxiang Hu, Peter Bailis, Ion Stoica, Zhijie Deng, Alvin Cheung, and Hao Zhang.
\newblock Online speculative decoding.
\newblock \emph{arXiv preprint arXiv:2310.07177}, 2023.

\bibitem[Mamou et~al.(2024)Mamou, Pereg, Korat, Berchansky, Timor, Wasserblat, and Schwartz]{mamou2024dynamic}
Jonathan Mamou, Oren Pereg, Daniel Korat, Moshe Berchansky, Nadav Timor, Moshe Wasserblat, and Roy Schwartz.
\newblock Dynamic speculation lookahead accelerates speculative decoding of large language models.
\newblock \emph{arXiv preprint arXiv:2405.04304}, 2024.

\bibitem[Miao et~al.(2023)Miao, Oliaro, Zhang, Cheng, Wang, Wong, Zhu, Yang, Shi, Shi, Chen, Arfeen, Abhyankar, and Jia]{miao2023specinfer}
Xupeng Miao, Gabriele Oliaro, Zhihao Zhang, Xinhao Cheng, Zeyu Wang, Rae Ying~Yee Wong, Alan Zhu, Lijie Yang, Xiaoxiang Shi, Chunan Shi, Zhuoming Chen, Daiyaan Arfeen, Reyna Abhyankar, and Zhihao Jia.
\newblock Specinfer: Accelerating generative large language model serving with speculative inference and token tree verification, 2023.

\bibitem[Monea et~al.(2023)Monea, Joulin, and Grave]{monea2023pass}
Giovanni Monea, Armand Joulin, and Edouard Grave.
\newblock Pass: Parallel speculative sampling.
\newblock \emph{arXiv preprint arXiv:2311.13581}, 2023.

\bibitem[Pope et~al.(2023)Pope, Douglas, Chowdhery, Devlin, Bradbury, Heek, Xiao, Agrawal, and Dean]{pope2023efficiently}
Reiner Pope, Sholto Douglas, Aakanksha Chowdhery, Jacob Devlin, James Bradbury, Jonathan Heek, Kefan Xiao, Shivani Agrawal, and Jeff Dean.
\newblock Efficiently scaling transformer inference.
\newblock \emph{Proceedings of Machine Learning and Systems}, 5, 2023.

\bibitem[Sadhukhan et~al.(2025)Sadhukhan, Chen, Chen, Tiwari, Lai, Shi, Yen, May, Chen, and Chen]{sadhukhan2025magicdec}
Ranajoy Sadhukhan, Jian Chen, Zhuoming Chen, Vashisth Tiwari, Ruihang Lai, Jinyuan Shi, Ian En-Hsu Yen, Avner May, Tianqi Chen, and Beidi Chen.
\newblock Magicdec: Breaking the latency-throughput tradeoff for long context generation with speculative decoding.
\newblock In \emph{The Thirteenth International Conference on Learning Representations}, 2025.
\newblock URL \url{https://openreview.net/forum?id=CS2JWaziYr}.

\bibitem[Santilli et~al.(2023)Santilli, Severino, Postolache, Maiorca, Mancusi, Marin, and Rodola]{santilli-etal-2023-accelerating}
Andrea Santilli, Silvio Severino, Emilian Postolache, Valentino Maiorca, Michele Mancusi, Riccardo Marin, and Emanuele Rodola.
\newblock Accelerating transformer inference for translation via parallel decoding.
\newblock In Anna Rogers, Jordan Boyd-Graber, and Naoaki Okazaki (eds.), \emph{Proceedings of the 61st Annual Meeting of the Association for Computational Linguistics (Volume 1: Long Papers)}, pp.\  12336--12355, Toronto, Canada, July 2023. Association for Computational Linguistics.
\newblock \doi{10.18653/v1/2023.acl-long.689}.
\newblock URL \url{https://aclanthology.org/2023.acl-long.689}.

\bibitem[Spector \& Re(2023)Spector and Re]{spector2023accelerating}
Benjamin~Frederick Spector and Christopher Re.
\newblock Accelerating {LLM} inference with staged speculative decoding.
\newblock In \emph{Workshop on Efficient Systems for Foundation Models @ ICML2023}, 2023.
\newblock URL \url{https://openreview.net/forum?id=RKHF3VYjLK}.

\bibitem[Stern et~al.(2018)Stern, Shazeer, and Uszkoreit]{stern2018blockwise}
Mitchell Stern, Noam Shazeer, and Jakob Uszkoreit.
\newblock Blockwise parallel decoding for deep autoregressive models.
\newblock \emph{Advances in Neural Information Processing Systems}, 31, 2018.

\bibitem[Su et~al.(2023)Su, Giannoula, and Pekhimenko]{su2023synergy}
Qidong Su, Christina Giannoula, and Gennady Pekhimenko.
\newblock The synergy of speculative decoding and batching in serving large language models.
\newblock \emph{arXiv preprint arXiv:2310.18813}, 2023.

\bibitem[Sun et~al.(2024{\natexlab{a}})Sun, Chen, Yang, Tian, and Chen]{sun2024triforce}
Hanshi Sun, Zhuoming Chen, Xinyu Yang, Yuandong Tian, and Beidi Chen.
\newblock Triforce: Lossless acceleration of long sequence generation with hierarchical speculative decoding.
\newblock \emph{arXiv preprint arXiv:2404.11912}, 2024{\natexlab{a}}.

\bibitem[Sun et~al.(2024{\natexlab{b}})Sun, Suresh, Ro, Beirami, Jain, and Yu]{sun2024spectr}
Ziteng Sun, Ananda~Theertha Suresh, Jae~Hun Ro, Ahmad Beirami, Himanshu Jain, and Felix Yu.
\newblock Spectr: Fast speculative decoding via optimal transport.
\newblock \emph{Advances in Neural Information Processing Systems}, 36, 2024{\natexlab{b}}.

\bibitem[Taori et~al.(2023)Taori, Gulrajani, Zhang, Dubois, Li, Guestrin, Liang, and Hashimoto]{alpaca}
Rohan Taori, Ishaan Gulrajani, Tianyi Zhang, Yann Dubois, Xuechen Li, Carlos Guestrin, Percy Liang, and Tatsunori~B. Hashimoto.
\newblock Stanford alpaca: An instruction-following llama model.
\newblock \url{https://github.com/tatsu-lab/stanford_alpaca}, 2023.

\bibitem[Team et~al.(2023)Team, Anil, Borgeaud, Wu, Alayrac, Yu, Soricut, Schalkwyk, Dai, Hauth, et~al.]{team2023gemini}
Gemini Team, Rohan Anil, Sebastian Borgeaud, Yonghui Wu, Jean-Baptiste Alayrac, Jiahui Yu, Radu Soricut, Johan Schalkwyk, Andrew~M Dai, Anja Hauth, et~al.
\newblock Gemini: a family of highly capable multimodal models.
\newblock \emph{arXiv preprint arXiv:2312.11805}, 2023.

\bibitem[Team et~al.(2024{\natexlab{a}})Team, Mesnard, Hardin, Dadashi, Bhupatiraju, Pathak, Sifre, Rivi{\`e}re, Kale, Love, et~al.]{team2024gemma}
Gemma Team, Thomas Mesnard, Cassidy Hardin, Robert Dadashi, Surya Bhupatiraju, Shreya Pathak, Laurent Sifre, Morgane Rivi{\`e}re, Mihir~Sanjay Kale, Juliette Love, et~al.
\newblock Gemma: Open models based on gemini research and technology.
\newblock \emph{arXiv preprint arXiv:2403.08295}, 2024{\natexlab{a}}.

\bibitem[Team et~al.(2024{\natexlab{b}})Team, Riviere, Pathak, Sessa, Hardin, Bhupatiraju, Hussenot, Mesnard, Shahriari, Ram{\'e}, et~al.]{team2024gemma2}
Gemma Team, Morgane Riviere, Shreya Pathak, Pier~Giuseppe Sessa, Cassidy Hardin, Surya Bhupatiraju, L{\'e}onard Hussenot, Thomas Mesnard, Bobak Shahriari, Alexandre Ram{\'e}, et~al.
\newblock Gemma 2: Improving open language models at a practical size.
\newblock \emph{arXiv preprint arXiv:2408.00118}, 2024{\natexlab{b}}.

\bibitem[Touvron et~al.(2023{\natexlab{a}})Touvron, Lavril, Izacard, Martinet, Lachaux, Lacroix, Rozi{\`e}re, Goyal, Hambro, Azhar, et~al.]{touvron2023llama}
Hugo Touvron, Thibaut Lavril, Gautier Izacard, Xavier Martinet, Marie-Anne Lachaux, Timoth{\'e}e Lacroix, Baptiste Rozi{\`e}re, Naman Goyal, Eric Hambro, Faisal Azhar, et~al.
\newblock Llama: Open and efficient foundation language models.
\newblock \emph{arXiv preprint arXiv:2302.13971}, 2023{\natexlab{a}}.

\bibitem[Touvron et~al.(2023{\natexlab{b}})Touvron, Martin, Stone, Albert, Almahairi, Babaei, Bashlykov, Batra, Bhargava, Bhosale, et~al.]{touvron2023llama2}
Hugo Touvron, Louis Martin, Kevin Stone, Peter Albert, Amjad Almahairi, Yasmine Babaei, Nikolay Bashlykov, Soumya Batra, Prajjwal Bhargava, Shruti Bhosale, et~al.
\newblock Llama 2: Open foundation and fine-tuned chat models.
\newblock \emph{arXiv preprint arXiv:2307.09288}, 2023{\natexlab{b}}.

\bibitem[Vaswani et~al.(2017)Vaswani, Shazeer, Parmar, Uszkoreit, Jones, Gomez, Kaiser, and Polosukhin]{vaswani2017attention}
Ashish Vaswani, Noam Shazeer, Niki Parmar, Jakob Uszkoreit, Llion Jones, Aidan~N Gomez, {\L}ukasz Kaiser, and Illia Polosukhin.
\newblock Attention is all you need.
\newblock \emph{Advances in neural information processing systems}, 30, 2017.

\bibitem[Wang et~al.(2023)Wang, Kordi, Mishra, Liu, Smith, Khashabi, and Hajishirzi]{wang-etal-2023-self-instruct}
Yizhong Wang, Yeganeh Kordi, Swaroop Mishra, Alisa Liu, Noah~A. Smith, Daniel Khashabi, and Hannaneh Hajishirzi.
\newblock Self-instruct: Aligning language models with self-generated instructions.
\newblock In Anna Rogers, Jordan Boyd-Graber, and Naoaki Okazaki (eds.), \emph{Proceedings of the 61st Annual Meeting of the Association for Computational Linguistics (Volume 1: Long Papers)}, pp.\  13484--13508, Toronto, Canada, July 2023. Association for Computational Linguistics.
\newblock \doi{10.18653/v1/2023.acl-long.754}.
\newblock URL \url{https://aclanthology.org/2023.acl-long.754}.

\bibitem[Xia et~al.(2024)Xia, Yang, Dong, Wang, Li, Ge, Liu, Li, and Sui]{xia2024unlocking}
Heming Xia, Zhe Yang, Qingxiu Dong, Peiyi Wang, Yongqi Li, Tao Ge, Tianyu Liu, Wenjie Li, and Zhifang Sui.
\newblock Unlocking efficiency in large language model inference: A comprehensive survey of speculative decoding.
\newblock \emph{arXiv preprint arXiv:2401.07851}, 2024.

\bibitem[Xu et~al.(2023)Xu, Yin, Jin, Zhang, Wei, Xu, and Liu]{xu2023llmcad}
Daliang Xu, Wangsong Yin, Xin Jin, Ying Zhang, Shiyun Wei, Mengwei Xu, and Xuanzhe Liu.
\newblock Llmcad: Fast and scalable on-device large language model inference.
\newblock \emph{arXiv preprint arXiv:2309.04255}, 2023.

\bibitem[Yang et~al.(2023{\natexlab{a}})Yang, Ge, Wang, Jiao, Jiang, Yang, Majumder, and Wei]{yang2023inference}
Nan Yang, Tao Ge, Liang Wang, Binxing Jiao, Daxin Jiang, Linjun Yang, Rangan Majumder, and Furu Wei.
\newblock Inference with reference: Lossless acceleration of large language models.
\newblock \emph{arXiv preprint arXiv:2304.04487}, 2023{\natexlab{a}}.

\bibitem[Yang et~al.(2024)Yang, Huang, Dai, and Chen]{yang2024multi}
Sen Yang, Shujian Huang, Xinyu Dai, and Jiajun Chen.
\newblock Multi-candidate speculative decoding.
\newblock \emph{arXiv preprint arXiv:2401.06706}, 2024.

\bibitem[Yang et~al.(2023{\natexlab{b}})Yang, Lee, Cho, Papailiopoulos, and Lee]{yang2023predictive}
Seongjun Yang, Gibbeum Lee, Jaewoong Cho, Dimitris Papailiopoulos, and Kangwook Lee.
\newblock Predictive pipelined decoding: A compute-latency trade-off for exact llm decoding.
\newblock \emph{arXiv preprint arXiv:2307.05908}, 2023{\natexlab{b}}.

\bibitem[Zhang et~al.(2024)Zhang, Wang, Wang, Zhang, and Cheng]{zhang2024recurrent}
Aonan Zhang, Chong Wang, Yi~Wang, Xuanyu Zhang, and Yunfei Cheng.
\newblock Recurrent drafter for fast speculative decoding in large language models.
\newblock \emph{arXiv preprint arXiv:2403.09919}, 2024.

\bibitem[Zhao et~al.(2024)Zhao, Huang, Han, Xiao, Liu, and Sun]{zhao2024ouroboros}
Weilin Zhao, Yuxiang Huang, Xu~Han, Chaojun Xiao, Zhiyuan Liu, and Maosong Sun.
\newblock Ouroboros: Speculative decoding with large model enhanced drafting.
\newblock \emph{arXiv preprint arXiv:2402.13720}, 2024.

\bibitem[Zhong et~al.(2024)Zhong, Yang, Li, Gong, Wang, and Huang]{zhong2024propd}
Shuzhang Zhong, Zebin Yang, Meng Li, Ruihao Gong, Runsheng Wang, and Ru~Huang.
\newblock Propd: Dynamic token tree pruning and generation for llm parallel decoding.
\newblock \emph{arXiv preprint arXiv:2402.13485}, 2024.

\bibitem[Zhou et~al.(2024)Zhou, Lyu, Rawat, Menon, Rostamizadeh, Kumar, Kagy, and Agarwal]{zhou2024distillspec}
Yongchao Zhou, Kaifeng Lyu, Ankit~Singh Rawat, Aditya~Krishna Menon, Afshin Rostamizadeh, Sanjiv Kumar, Jean-Fran{\c{c}}ois Kagy, and Rishabh Agarwal.
\newblock Distillspec: Improving speculative decoding via knowledge distillation.
\newblock In \emph{The Twelfth International Conference on Learning Representations}, 2024.
\newblock URL \url{https://openreview.net/forum?id=rsY6J3ZaTF}.

\end{thebibliography}
\bibliographystyle{colm2025_conference}

\newpage
\appendix
\section{Limitations}
\label{sec:limitation}

Our theoretical result contains a problem-specific constant $\Delta$ which is hard to analyze theoretically or estimate empirically. Nevertheless, the choice of the stopping threshold $h$ can be determined through hyperparameter search; see Appendix~\ref{appendix:ablation}.
As is the case with all speculative decoding algorithms, our method relies on the implicit assumption that the draft model and the target model align well. For a weak draft model, the acceptance prediction head may perform badly.

\section{Additional Related Work}
\label{appendix:related}

Large language models are mostly based on Transformer architectures~\citep{vaswani2017attention} that auto-regressively predict the probability of the next token given its predecessors.
One bottleneck of the inference speed lies in the fact that auto-regressive decoding is an inherently non-parallelizable sequential operation: the probabilities of future tokens depend on the current token and there is no trivial way to skip the current token when predicting future tokens.
Therefore, the inference time of auto-regressive decoding scales linearly with the number of the generated tokens.

However, the time of a forward pass to compute the log probabilities of the tokens through transformers is nearly constant for batched sequences with different lengths within a proper range, thanks to the increasingly powerful parallel computing units~\citep{pope2023efficiently, vaswani2017attention, chen2023accelerating, pmlr-v202-leviathan23a}.

Therefore, to overcome the bottleneck of the auto-regressive decoding, one can find a fast way to generate $K$ tokens, which often increases FLOPs, and the ask the target model to verify and correct the candidates~\citep{stern2018blockwise, chen2023accelerating, pmlr-v202-leviathan23a}; see a comprehensive survey~\citep{xia2024unlocking}. 
For those methods to work, we assume that we have enough computational resources (e.g. CUDA memories) to support the increased concurrency. Nevertheless, in the long-context generation regime, the memory issue becomes prominent, which requires additional KV-cache management techniques such as compression or retrieval~\citep{li2024snapkv, sun2024triforce}.

\textbf{Improvements of Speculative Decoding Methods}

The performance of speculative decoding depends on how well the draft model aligns with the target model, and how fast the draft model is compared to the target model. People have been improving speculative decoding in two aspects: (1) making the draft model align better with the target model via distillation~\citep{zhou2024distillspec, agarwal2024policy} and online learning~\citep{liu2023online};
and (2) making the token generation faster and cheaper, e.g. training multiple smaller draft models from stratch~ \citep{miao2023specinfer}. 

In addition, the candidate tokens can be generated without a separate draft model~\citep{stern2018blockwise, li2024eagle, du2024glide, bhendawade2024speculative}, such as building additional modules that predict the next $k$ tokens (Medusa heads~\citep{cai2024medusa}, RNN heads~\citep{zhang2024recurrent}, soft tokens~\citep{monea2023pass}), early-exiting methods that reuse the intermediate representations of the target model~\citep{liu2024kangaroo,yang2023predictive,bae2023fast}, and retrieval-based methods that involve constructing an $n$-gram datastore and using retrieval to generate candidates~\citep{he2023rest, zhao2024ouroboros, yang2023inference, fu2024break}. 

Those techniques can be combined, resulting in a heirachical system~\citep{spector2023accelerating,zhao2024ouroboros, sun2024triforce}.

\textbf{Token Tree Generation, Verification and Pruning.}

Paralleling across the batch dimension via token trees is another direction to increase throughputs~\citep{miao2023specinfer, xu2023llmcad, su2023synergy}. 
For greedy decoding, token tree generation and verification are studied in~\citep{cai2024medusa}. For the stochastic sampling setting, REST~\citep{he2023rest} proposes a straightforward approach: keeping the token paths that coincide with the stochastic tokens given by the target model. 
There are also researches extending the stochastic speculative decoding to the token tree setting, which often needs to adjust the drafting and verification probabilities to ensure unbiasedness, e.g. MCSD~\citep{yang2024multi}, Recursive SD~\citep{jeon2024recursive}, Sequoia~\citep{chen2024sequoia}, EAGLE~\citep{li2024eagle}, SpecTR~\citep{sun2024spectr}.

One important problem to study is how to construct and prune the token tree to maximize throughputs and avoid heavy communication overheads, which is studied in~\citep{chen2024sequoia, zhong2024propd}.
Our work can serve as a starting point towards the problem, as the candidate length $K$ can be viewed as the depth of a token tree with only one branch.

\textbf{Diffusion language models.}  Diffusion language models either in the discrete space (see D3PM~\citep{austin2021structured} and its follow-ups) or in the embedding space (see Diffusion-LM~\citep{li2022diffusion} and its follow-ups) are non-autoregressive language models, whose generation time can scale sub-linearly with the sequence length. 
BERT-type encoder-only models and auto-regressive decoder-only models can be also viewed as diffusion model, with mask prediction and next-token prediction being the denoising operation~\citep{austin2021structured}.  
Viewing next-token prediction as \textit{Jacobi iteration}~\citep{santilli-etal-2023-accelerating} and \textit{denoising operation} is a powerful idea and it leads to subsequent work such as lookahead decoding~\citep{fu2024break} and consistency LLMs~\citep{kou2024cllms}.

\subsection{Discussion: Sub-optimality of Heuristic Methods }
\label{appendix:discussion}

In this section, we discuss the potential sub-optimality of several training-free heuristic methods for determining the candidate lengths. For example, it may be tempting to use the entropy of the draft model $\mathcal{H}(q(\cdot \mid x_{\text{prefix}}, Y_1,\dots, Y_{i-1}))$ or merely the likelihood of the sampled draft candidate token $q(Y_i \mid x_{\text{prefix}}, Y_1,\dots, Y_{i-1}))$ as a surrogate for the acceptance probability, and choose to stop the current round of speculation when these indicators fall under a threshold. However, we point out that this type of training-free heuristics confuses the inherent uncertainty of the draft distribution with the alignment between draft and target distributions, which is fundamentally flawed. 

First of all, the theoretical acceptance probability is $\min \Big(1, \frac{p(Y_i|x_{\text{prefix}}, Y_1,\dots,Y_{i-1})}{q(Y_i|x_{\text{prefix}}, Y_1,\dots,Y_{i-1})}\Big)$, which depends on the target distribution $p$. Therefore, heuristic methods that do not incorporate the information of the target model will be sub-optimal. 

Furthermore, when the draft model and the target model align well, the acceptance probability will be high regardless of the entropy of the current draft candidate token or the likelihood of the sampled $Y_i$. For the extreme case when the draft model and the target model aligns perfectly well i.e.,  $p(Y_i|x_{\text{prefix}}, Y_1,\dots,Y_{i-1})  = q(Y_i|x_{\text{prefix}}, Y_1,\dots,Y_{i-1})$, by the rejection sampling scheme of speculative decoding, the sampled candidate tokens are \textit{guaranteed} to be accepted, and the optimal candidate lengths will be infinite. Using either entropy-based methods or likelihood-based methods will stop the speculation earlier and results in sub-optimal performance. 

\section{Additional Background on Speculative Decoding Algorithm}
\label{appendix:algo}

\textbf{Rejection Sampling.} The algorithmic foundation of the Speculative Decoding algorithm lies in rejection sampling.
Specifically, if we want to sample from a target discrete distribution $p(x)$, we first sample $x$ from a draft distribution $q(x)$. We accept the sample $x$ with probability $\min(1, \frac{p(x)}{q(x)})$; otherwise we replace it with a sample from the modified distribution $\mathrm{Norm}[(p-q)_+]$, where $z_+ = \max(z, 0)$ is the positive part of $z$ and $\mathrm{Norm}[f] = \frac{f(\cdot)}{\sum_x f(x)}$ normalizes a function $f$ to make it a proper probability distribution. The proof of the unbiasedness of rejection sampling can be found in \citet{chen2023accelerating}.

\textbf{Speculative Decoding.} Speculative decoding extends to the auto-regressive generation scenarios by chaining $K$ rejection sampling procedures together. The full algorithm is provided in Algorithm~\ref{alg:sd}.

\begin{algorithm} %
	\caption{Speculative Decoding~\citep{chen2023accelerating, pmlr-v202-leviathan23a}} %
	\label{alg:sd} %
	\begin{algorithmic} %
	\REQUIRE draft model $q$, target model $p$, prefix $x_{\text{prefix}}$, number of candidate tokens $K$.
        \FOR{$i=1$ to $K$ }
            \STATE{Compute $q_i = q(\cdot \mid x_{\text{prefix}}, y_1, \dots, y_{i-1})$.}
            \STATE{Sample $y_i \sim q_i$.}
        \ENDFOR
        \STATE{ Compute \textit{in parallel} $p_i = p(\cdot \mid x_{\text{prefix}}, y_1, \dots, y_{i-1})$ for $i=1,\dots,K+1$. }
        \STATE{ Sample $r_1,\dots,r_K$ with $r_i \sim \mathrm{Unif}[0,1]$, $i=1,\dots,K$. }
        \STATE{ Compute the number of accepted tokens $n=\min\Big (\{i-1 \mid r_i \geq p_i(y_i)/q_i(y_i)\} \cup K \Big) $.}
	\IF{$n < K$}
            \STATE{Sample $y'$ from the modified distribution $\mathrm{Norm}[(p_{n+1} - q_{n+1})_+]$ }
        \ELSE
            \STATE{Sample $y'$ from $p_{K+1}$}
        \ENDIF
        \STATE{\textbf{Return} $x_{\text{prefix}}, y_1,\dots,y_{n}, y'$}
	\end{algorithmic}
\end{algorithm}

\section{Additional Experimental Results}
\label{appendix:exp}
\subsection{Experimental Setups}
\label{appendix:exp:setup}
The subsection continues Section~\ref{sec:exp:setup}.

\textbf{Datasets.} We adopt three datasets in our experiments: (1) Alpaca~\citep{alpaca}, an instruction-following dataset generated using Self-Instruct~\citep{wang-etal-2023-self-instruct} from OpenAI's \texttt{text-davinci-003} model; (2) HumanEval~\citep{chen2021evaluating}, a test dataset containing Python code synthesis problems; and (3) GSM8K~\citep{cobbe2021training}, a dataset of high-school math problems. We only use prompts of the datasets and do not use responses. 

\textbf{Dataset splits.}
We split the Alpaca dataset into train/dev/test splits, containing 40k, 10k, 2k prompts, respectively. We use train split to train the prediction heads and evaluate them on the dev split. We benchmark the performance of \ours on the test split.
For HumanEval and GSM8K, we only use them for benchmarking the out-of-distribution (OOD) performance of \ours. 
For each test dataset, we subsample $150$ examples for benchmarking the performances.

\textbf{Mixing probability.} As in Section~\ref{sec:method:train}, we mix the response tokens from the generations from the target model and the predicted next-tokens from the draft model. We set an aggressive value $r\%=15\%$ so only $15\%$ of the tokens are from the target model, as we find empirically that the draft model and the target model often align well. Setting a smaller $r$ increases the training efficiency as more supervision signals are used.

\textbf{Training Details.} We train all the acceptance prediction heads on the train split of the Alpaca dataset for 3 epochs with batch size 32. We use Adam optimizer and a cosine learning rate schedule with the initial learning rate $5e-5$.

\textbf{Hardware configuration.} We use 2 NVIDIA A100 GPUs with 80G memory for the experiments. We shard the 70B model across the two devices and communication overhead occurs when inferring with llama-2-chat 70B. When doing speculative decoding, the 7B model is loaded only on one device.

\textbf{Inference setting.} We set the maximal sequence length to be $512$. We use temperature $T=1$ and adopt top-k sampling with $k=50$.
We do not integrate KV cache management techniques such as PagedAttention~\citep{kwon2023efficient} or KV cache pre-allocation.

\textbf{Experiments Compute Resources.}  The required compute resources are estimated to be 500 hours on 2 NVIDIA A100-80G GPUs for the training dataset generation,  400 hours on 1 NVIDIA A100-80G GPU for training 20 acceptance prediction heads (sweeping $D$ from $0$ to $4$ and $w_\text{rej}$ among $1,3,6,12$), 500 hours on 2 NVIDIA A100-80G GPUs for the whole evaluation set. The full research project would require at least 2x the reported compute, as there were preliminary experiments that are not in the paper.

\subsection{Forward Time Analysis}
\label{appendix:time}
\begin{figure}[h]
    \centering
    \includegraphics[width=0.45\textwidth]{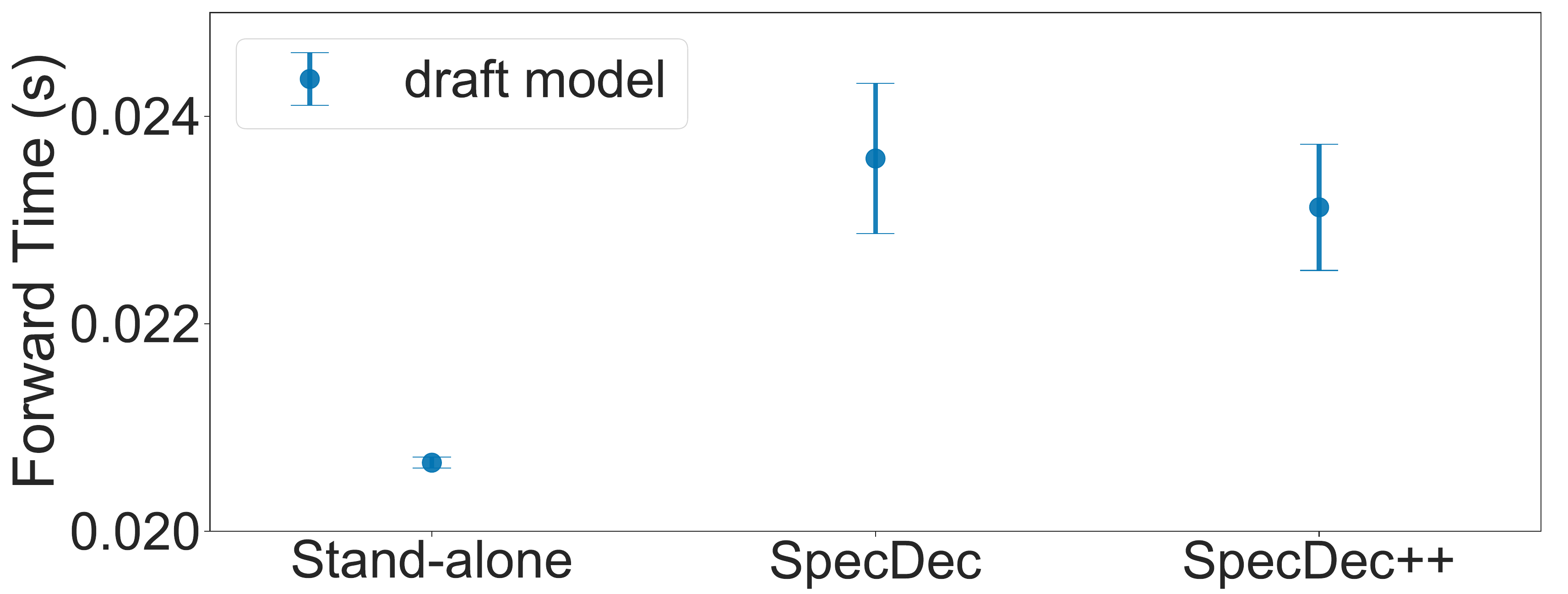}
    \includegraphics[width=0.45\textwidth]{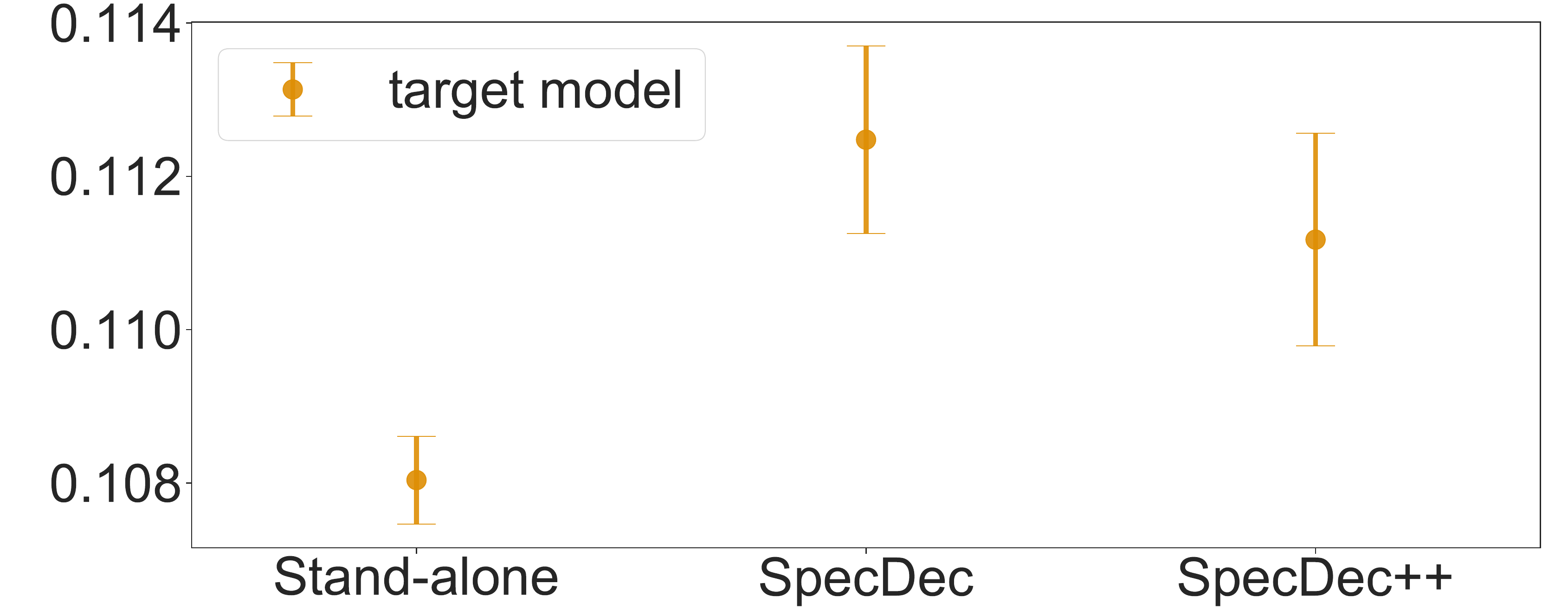}
    \caption{The forward time of the draft model (llama-2-chat-7B) and the target model (llama-2-chat-70B) under different settings. 
    For each setting, we perform linear regression to calculate the forward times and then average them across different datasets. 
    The additional cost of the acceptance prediction head is negligible compared to the systematic error and the random noise of the environment. 
    Full results are in Table~\ref{tab:time}.
    } 
    \label{fig:time}
\end{figure}

We report the full results of the linear regression in Section~\ref{sec:time} in Table~\ref{tab:time}. We also visualize $t_\text{draft}$ and $t_\text{target}$ across the three settings in Figure~\ref{fig:time}.

\begin{table}[htbp]
    \caption{The forward time of the draft model (llama-2-chat-7B) and the target model (llama-2-chat-70B) under different settings and different datasets. We perform linear regression to calculate the forward times.}
    \label{tab:time}
    \centering
    \begin{tabular}{llccc}
    \hline
    Setting                     & Dataset   & $t_{\text{draft}}$ & $t_{\text{target}}$ & $R^2$ \\ \hline
    \multirow{4}{*}{stand-alone} & Alpaca    & 0.0206 & 0.108 & 0.9994 \& 0.9998  \\
                                & HumanEval & 0.0207 & 0.107 & 0.9994 \& 0.9998 \\
                                & GSM8K     & 0.0206 &  0.109 & 0.9990 \& 0.9992   \\ \cline{2-5} 
                                & average   & 0.0207 $\pm$ 0.0001 & 0.108 $\pm$ 0.001 &    \\ \hline
    \multirow{4}{*}{SpecDec}    & Alpaca    & 0.0232 & 0.114 & 0.9983  \\
                                & HumanEval & 0.0246 & 0.111 & 0.9965  \\
                                & GSM8K     & 0.0229 & 0.113 & 0.9926 \\ \cline{2-5} 
                                & average   & 0.0236 $\pm$ 0.0007 & 0.112 $\pm$ 0.001 &    \\ \hline
    \multirow{4}{*}{SpecDec++}  & Alpaca    & 0.0240 & 0.110 & 0.9982  \\
                                & HumanEval & 0.0229 & 0.111 &  0.9880 \\
                                & GSM8K     & 0.0225 & 0.113 & 0.9925  \\ \cline{2-5} 
                                & average   & 0.0231 $\pm$ 0.0006 & 0.111 $\pm$ 0.001 &    \\ \hline
    \end{tabular}
\end{table}

From Figure~\ref{fig:time}, we see that the additional cost of the acceptance prediction head is negligible. Besides, in the standalone setting where only the draft model or the target model is used, both $t_{\text{draft}}$ and $t_{\text{target}}$ decrease, which indicates that speculative decoding induces minor overhead in forward passes. 

After getting $t_{\text{draft}}$ and $t_{\text{target}}$, we use Equation~\eqref{eq:costfn2} to calculate the theoretical throughputs (tokens per second), which match the noisier empirical throughputs well with relative error $\leq 6.2\%$ for all prompts. 

\subsection{Ablation Studies.} 
\label{appendix:ablation}
We fix $w_{\text{acc}} = 1$ and study how the hyperparameters $w_{\text{rej}}, D, h$ influence the final throughputs (tokens per second). First, we calculate the (unweighted) binary KL divergence between the ground-truth probability and the predicted probability, i.e., 
\[
   \mathrm{KL}( p || q) = p\log \frac{p}{q} + (1-p) \log \frac{1-p}{1-q}.
\]
As $\mathrm{KL}( p || q)  = \mathrm{BCE}(p||q) - H(p)$, the binary KL divergence is a metric for how well the acceptance prediction head fits the ground-truth probabilities.
Next, for each acceptance prediction head, we report the best throughput by varying the stopping threshold $h$ among $\{0.1,0.3,0.5,0.7,0.9\}$, and the corresponding $h$ that achieves the best performance. The results are summarized in Table~\ref{tab:head}.

From the table, we see that increasing $w_{\text{rej}}$ increases the \textit{unweighted} eval KL. Most of the prediction heads trained with $w_{\text{rej}} = 1$ perform the best with $h = 0.3$ under all three datasets, and similarly, most prediction heads trained with $w_{\text{rej}} = 3, 6, 12$ perform the best with $h=0.5, 0.7, 0.9$, respectively. This synergy between $w_{\text{rej}} = 1$ and $h$ is expected, since increasing $w_{\text{rej}} = 1$ forces the acceptance prediction head to focus more on the cases where the candidate token is rejected and thus mitigates the over-confidence issue. In return, the stopping threshold $h$ can be set to a higher value to adjust for the increased predicted probability of existing one rejection.

We bold the throughputs that are above $99\%$ of the maximum throughput of the same dataset. We see that there are two sets of hyperparameters that consistently achieve $99\%$ of the maximum throughputs across the three datasets: $w_{\text{rej}} = 6 $, $D = 3$, $h=0.7$ and $w_{\text{rej}} = 6 $, $D = 4$, $h=0.7$.

\begin{table}[htbp]
    \caption{The performance of the acceptance prediction heads with different loss weights $w_\text{rej}$ and network depths $D$. The train/eval KL refers to the binary KL divergence between the ground-truth probability and the predicted probability. For the three datasets, we report the best throughput and the corresponding stopping threshold $h$. The throughputs are \textbf{bolded} if they are above $99\%$ of the maximum throughput of the same dataset. }
    \label{tab:head}
    \centering
    \begin{tabular}{llccccc}
    \hline
    $w_{\text{rej}}$      & Depth $D$   & train/KL & eval/KL & Alpaca  &  HumanEval  &  GSM8K \\ \hline
1 & 0 & 0.422 & 0.412 & 18.48 ($h=0.3$) & 19.91 ($h=0.5$) & 20.32 ($h=0.3$) \\
1 & 1 & 0.409 & 0.390 & 18.39 ($h=0.3$) & 20.29 ($h=0.3$) & 20.44 ($h=0.3$) \\
1 & 2 & 0.391 & 0.387 & \textbf{18.87} ($h=0.3$) & 20.26 ($h=0.3$) & \textbf{20.87} ($h=0.3$) \\
1 & 3 & 0.387 & 0.384 & \textbf{18.82} ($h=0.3$) & 20.10 ($h=0.3$) & \textbf{20.86} ($h=0.3$) \\
1 & 4 & 0.384 & 0.383 & 18.57 ($h=0.3$) & \textbf{20.51} ($h=0.3$) & 20.73 ($h=0.3$) \\ \hline
3 & 0 & 0.515 & 0.491 & 18.31 ($h=0.5$) & 20.12 ($h=0.7$) & 20.36 ($h=0.5$) \\
3 & 1 & 0.479 & 0.461 & \textbf{18.88} ($h=0.5$) & 20.32 ($h=0.5$) & 20.70 ($h=0.5$) \\
3 & 2 & 0.475 & 0.458 & 18.60 ($h=0.5$) & 20.17 ($h=0.5$) & 20.61 ($h=0.3$) \\
3 & 3 & 0.462 & 0.454 & \textbf{18.76} ($h=0.5$) & 20.32 ($h=0.5$) & \textbf{20.88} ($h=0.5$) \\
3 & 4 & 0.465 & 0.451 & \textbf{18.88} ($h=0.5$) & \textbf{20.50} ($h=0.7$) & \textbf{20.82} ($h=0.5$) \\ \hline
6 & 0 & 0.657 & 0.637 & 18.67 ($h=0.7$) & 19.90 ($h=0.9$) & 20.24 ($h=0.7$) \\
6 & 1 & 0.620 & 0.596 & \textbf{18.75} ($h=0.7$) & 20.09 ($h=0.9$) & \textbf{20.86} ($h=0.7$) \\
6 & 2 & 0.607 & 0.589 & 18.65 ($h=0.7$) & 20.17 ($h=0.9$) & 20.70 ($h=0.7$) \\
6 & 3 & 0.617 & 0.582 & \textbf{18.80} ($h=0.7$) & \textbf{20.47} ($h=0.7$) & \textbf{20.95} ($h=0.7$) \\
6 & 4 & 0.603 & 0.575 & \textbf{18.87} ($h=0.7$) & \textbf{20.61} ($h=0.7$) & \textbf{20.77} ($h=0.7$) \\ \hline
12 & 0 & 0.922 & 0.871 & 18.55 ($h=0.9$) & 19.93 ($h=0.9$) & 20.62 ($h=0.9$) \\
12 & 1 & 0.830 & 0.805 & \textbf{18.71} ($h=0.9$) & 20.25 ($h=0.9$) & 20.73 ($h=0.9$) \\
12 & 2 & 0.834 & 0.794 & 18.58 ($h=0.9$) & 20.39 ($h=0.9$) & \textbf{20.77} ($h=0.7$) \\
12 & 3 & 0.801 & 0.781 & \textbf{18.76} ($h=0.9$) & 20.29 ($h=0.9$) & 20.67 ($h=0.9$) \\
12 & 4 & 0.799 & 0.773 & \textbf{18.82} ($h=0.9$) & 20.19 ($h=0.9$) & 20.65 ($h=0.9$) \\ \hline
    \end{tabular}
\end{table}

\subsection{New Experiments on Gemma}
\label{appendix:gemma}

\begin{figure}[h]
\vspace{-1em}
    \centering
    \includegraphics[width=0.31\textwidth]{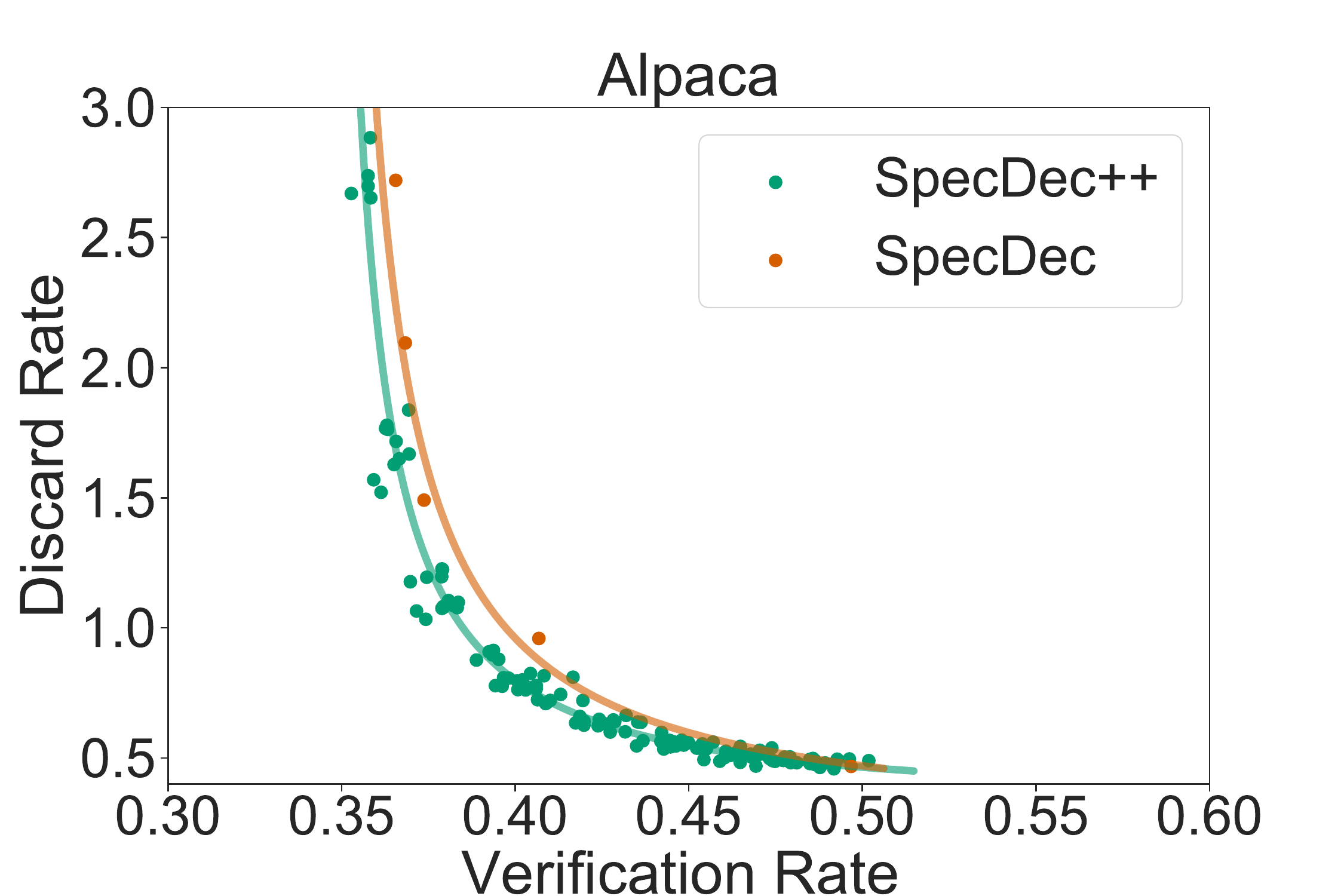}
    \includegraphics[width=0.31\textwidth]{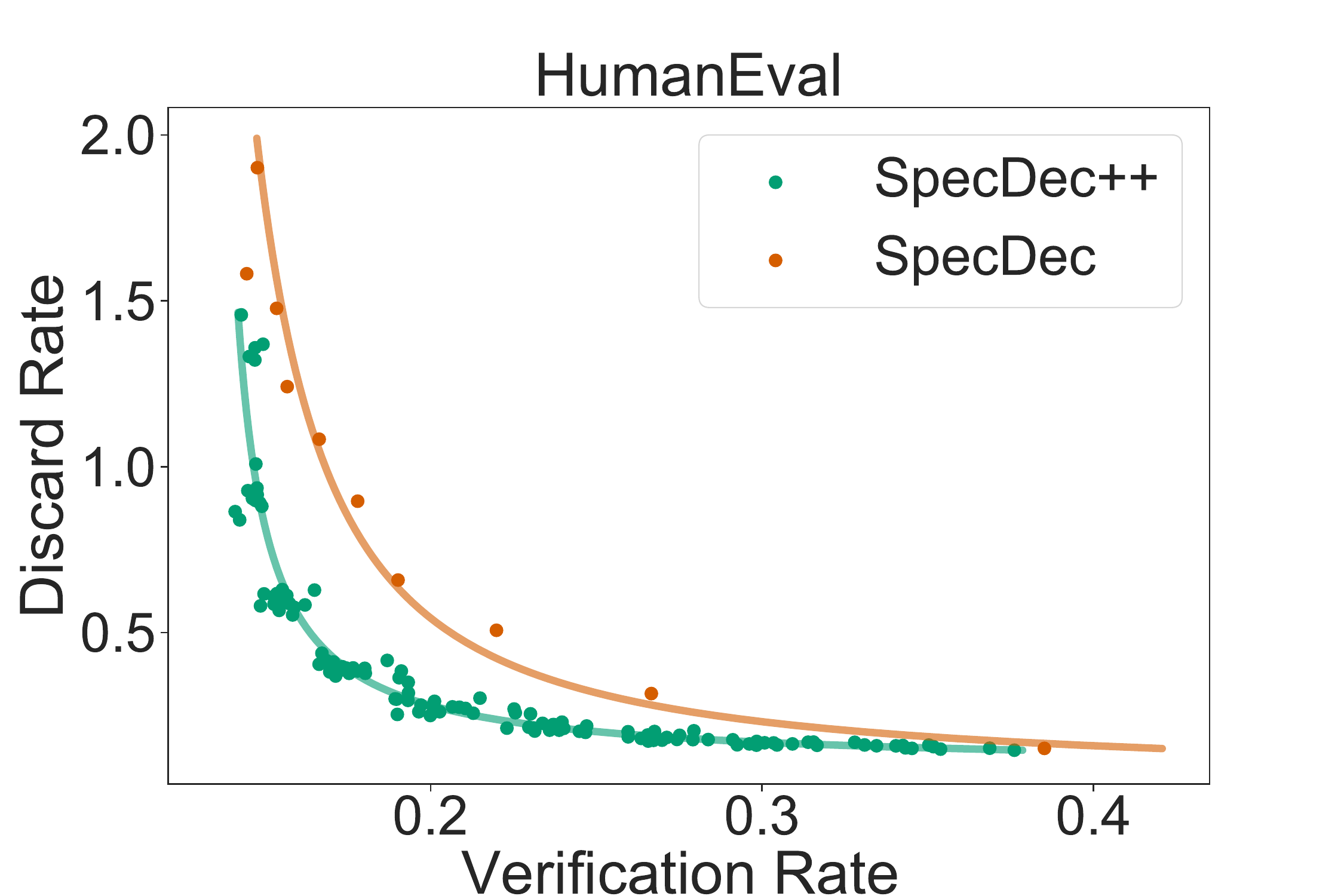}
    \includegraphics[width=0.31\textwidth]{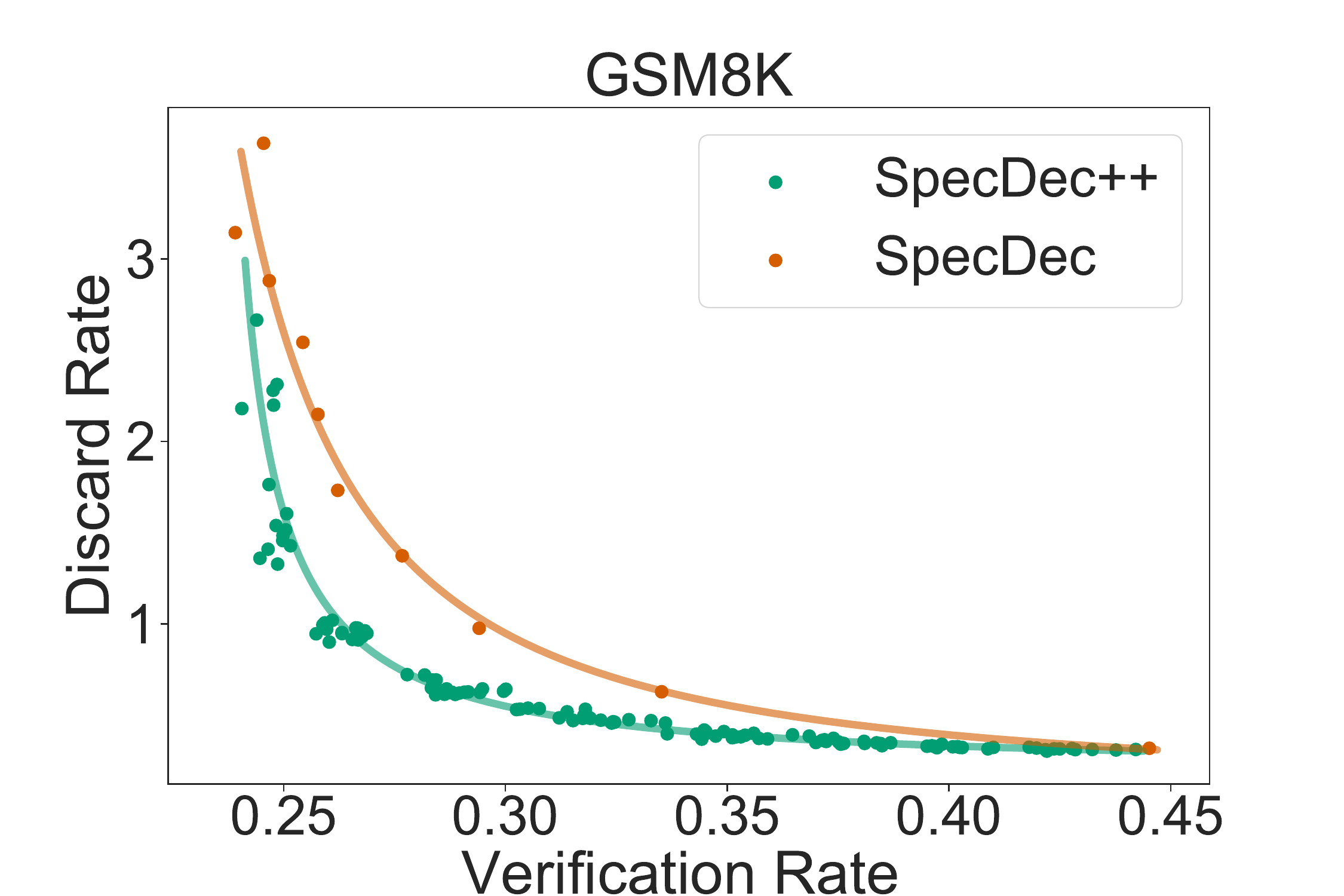}
    \caption{The average verification rates $N_{\text{target}}/N$ and the average discard rates $N_{\text{discarded}}/N$ for SpecDec with different candidate lengths and \ours with different acceptance prediction heads and stopping thresholds. \ours has better Pareto frontiers than SpecDec on both the in-distribution dataset Alpaca and the two out-of-distribution datasets HumanEval and GSM8K. The model pair is Gemma-1.1-2B-it/Gemma-2-27B-it.
    } 
    \label{fig:gemma:pareto}
\end{figure}

We further validate \ours by repeating all the experiments on a new pair of models: Gemma-1.1-2B-it~\citep{team2024gemma} and Gemma-2-27B-it~\citep{team2024gemma2}. The Pareto frontiers of verification rates versus discard rates are plotted in Figure~\ref{fig:gemma:pareto}, and the empirical speedup is reported in Table~\ref{tab:main:gemma:result}. Specifically, we see that 
\begin{itemize}
    \item SpecDec++ has better Pareto frontiers in terms of discard rates v.s. verification rates tradeoff than the baseline SpecDec algorithm on the Alpaca, GSM8K, and HumanEval datasets. The Pareto improvement indicates that SpecDec++ will have better speedups than SpecDec under arbitrary hardware configurations.
    \item When deployed on 1 Nvidia A100-80G GPU, SpecDec++ achieves a relative 1.4\%, 12.4\%, and 7.7\% improvement over the baseline methods on the Alpaca, HumanEval, and GSM8K datasets, respectively.
\end{itemize}

\begin{table}[h]
    \caption{The speedup achieved by \ours compared to the best speedup achieved by the speculative decoding baseline on Alpaca, HumanEval, and GSM8K datasets. The model pair is Gemma-1.1-2B-it/Gemma-2-27B-it.}
    \label{tab:main:gemma:result}
    \centering
    \begin{tabular}{llccc}
    \hline
     Dataset   &  Alpaca  & HumanEval & GSM8K  \\ \hline
    \ours & \textbf{1.35x} & \textbf{2.07x}   & \textbf{1.58x}  \\
      SpecDec (baseline)  & 1.33x  &  1.84x &  1.47x   \\ \hline
      relative improvement & +1.4\% & +12.4\% & +7.7\%
    \end{tabular}
\end{table}

We observe fewer improvements on the Alpaca dataset and the GSM8K dataset compared to the llama-2-7b-chat/llama-2-70b-chat model pair. This phenomenon may be caused by weaker alignment between the draft Gemma model and the target Gemma model.  On Alpaca, the baseline SpecDec achieves 1.33x speedup for the Gemma model pair but it achieves 1.90x speedup for the Llama model pair; on GSM8K, the baseline SpecDec achieves 1.47x speedup for the Gemma model pair but it achieves 2.07x speedup for the Llama model pair. We see our chosen Gemma model pair performs worse than the Llama model pair, although the model size ratios are roughly the same for the two model families (27B/2B v.s. 70B/7B).

We suspect that for the Gemma-1.1-2B-it and Gemma-2-27B-it model pair on the Alpaca dataset, our acceptance prediction head ends up suggesting an approximately fixed draft length, therefore only achieving a small improvement over the baseline. For settings like math (GSM8K) and coding (HumanEval), the generation involves a mixture of tokens of verbal reasoning and tokens of math calculation/code writing, which naturally require an adaptive draft length.

In summary, although the specific numbers of the improvement vary across different model pairs and different prompt settings, the proposed SpecDec++ indeed achieves an adequate amount of improvements over the baseline. Backed by the theoretical results, our proposed method is a principled extension of the SpecDec method, and adaptively determining the draft length naturally includes the baseline (a fixed draft length) as a special case.

\clearpage
\section{Theoretical Analysis}
\label{sec:theory}
In the section, we present the proof of Theorem~\ref{thm:main}.

For any time-homogeneous policy $\pi$, we define a random variable $C^\pi(s, a)$ as the total cost-to-go from the current state $s = (x_\text{prefix}, (Y_1,\dots, Y_k))$ when taking action $a$. 
\[
    C^\pi(s, a) = \sum_{i = 1}^{M}  c(s_i, a_i, s_{i+1}), \text{ with } s_1=s, a_1 = a,
\]
where the next state $s_{i+1}$ given $(s_i, a_i)$ follows the stochastic transition of the MDP, $a_i = \pi(s_i)$ for $i\geq 2$, and $M$ is a random variable of the number of total steps. We make the assumption that $\pi$ has an upper bound for the number of candidate tokens, so we exclude the cases where the policy $\pi$ potentially leads to an infinite loop and hence $M < \infty$. Let $C^\pi(s) = C^\pi(s, \pi(s))$. 

\begin{proof}[proof of Theorem~\ref{thm:main}]

We analyze the difference $C^\pi (s, \Continue) - C^\pi (s, \Stop)$ for three cases.

\textbf{Case 1.} $\mathcal{E}_1 = \{\exists 1 \leq i \leq k+1, \text{ such that } Y_i \text{ is rejected}\}$. 

Let $x_\text{prefix}'$ be the next prefix given by the speculative decoding algorithm, where the first rejected token among $(Y_1,\dots,Y_{k+1})$ is replaced by the token from the modified distribution. We know that 
\[
    C^\pi (s, \Stop) = c_1 + c_2 + C^\pi((x_\text{prefix}', \varnothing)).
\]
If we choose to $\Continue$ at the current step, we know that no matter how many additional steps we continue to generate draft tokens, we will eventually discard them and get the same new prefix $x_\text{prefix}'$. Let $N^\pi_{\Continue}(s)$ be the total number of extra $\Continue$'s induced by the policy $\pi$ given the current state $s$ and action $\Continue$. We have 
\[
    C^\pi (s, \Continue) = c_1 + c_1\cdot(1+N^\pi_{\Continue}(s)) + c_2 + C^\pi((x_\text{prefix}', \varnothing)).
\]
In summary, we have 
\[
    C^\pi (s, \Continue) - C^\pi (s, \Stop) \geq c_1, \text{ conditioned on } \mathcal{E}_1.
\]

\textbf{Case 2.} $\mathcal{E}_2 = \{\forall 1 \leq i \leq k+1, Y_i \text{ is accepted}, Y_{k+2} \text{ is rejected}\}$. 

If we stop the current round of speculation, then all the candidate tokens $(Y_1, \dots, Y_{k+1})$ will be accepted and an additional $X_{k+2}$ is sampled from $p(\cdot \mid x_{\text{prefix}}, Y_1, \dots, Y_{k+1})$.
\[
    C^\pi (s, \Stop) = c_2 + C^\pi (((x_{\text{prefix}}, Y_1, \dots, Y_{k+1}, X_{k+2}), \varnothing)).
\]
Again, if we choose to $\Continue$ at the current step, as $Y_{k+2}$ is rejected, future generated tokens beyond $Y_{k+2}$ will also be discarded. After the verification, $Y_{k+2}$ will be replaced by $W_{k+2} \sim \mathrm{Norm}[(p(\cdot|x_{\text{prefix}},Y_1\dots,Y_{k+1}) - q(\cdot|x_{\text{prefix}},Y_1\dots,Y_{k+1}))_+] $. Let $N^\pi_{\Continue}(s)$ be the total number of extra $\Continue$'s induced by the policy $\pi$ given the current state $s$ and action $\Continue$. We have 
\[
    C^\pi (s, \Continue) = c_1\cdot(1+N^\pi_{\Continue}(s)) + c_2 + C^\pi (((x_{\text{prefix}}, Y_1, \dots, Y_{k+1}, W_{k+2}), \varnothing)).
\]
Denote $\Delta_1 = C^\pi (((x_{\text{prefix}}, Y_1, \dots, Y_{k+1}, X_{k+2}), \varnothing)) - C^\pi (((x_{\text{prefix}}, Y_1, \dots, Y_{k+1}, W_{k+2}), \varnothing))$. In summary, we have 
\[
    C^\pi (s, \Continue) - C^\pi (s, \Stop) \geq c_1 - \Delta_1, \text{ conditioned on } \mathcal{E}_2.
\]

\textbf{Case 3.} $\mathcal{E}_3 = \{\forall 1 \leq i \leq k+2, Y_i \text{ is accepted}\}$. 

Similar to Case 2, if we stop the current round of speculation, then all the candidate tokens $(Y_1, \dots, Y_{k+1})$ will be accepted, and an additional $X_{k+2}$ is sampled from $p(\cdot \mid x_{\text{prefix}}, Y_1, \dots, Y_{k+1})$.
\[
    C^\pi (s, \Stop) = c_2 + C^\pi (((x_{\text{prefix}}, Y_1, \dots, Y_{k+1}, X_{k+2}), \varnothing)).
\]
If we choose to $\Continue$ at the current step, there is no immediate cost at the current step and we transit to $(x_{\text{prefix}}, (Y_1, \dots, Y_{k+1}))$.
\[
    C^\pi (s, \Continue) =  C^\pi ((x_{\text{prefix}}, (Y_1, \dots, Y_{k+1}))).
\]
Denote $\Delta_2 = C^\pi (((x_{\text{prefix}}, Y_1, \dots, Y_{k+1}, X_{k+2}), \varnothing)) - C^\pi ((x_{\text{prefix}}, (Y_1, \dots, Y_{k+1})))$. We have 
\[
    C^\pi (s, \Continue) - C^\pi (s, \Stop) \geq -c_2 - \Delta_2, \text{ conditioned on } \mathcal{E}_3.
\]

\textbf{Summary.} At the current state, the values of $(Y_1, \dots, Y_k)$ are known. We calculate the conditional expectation of $C^\pi (s, \Continue) - C^\pi (s, \Stop)$ given the current observation. For simplicity of notation, we do not explicitly write out the condition on $(Y_1, \dots, Y_k)$.
\begin{align*}
    & \EE[C^\pi (s, \Continue) - C^\pi (s, \Stop)] \\
     \geq & \PP(\mathcal{E}_1) c_1 +  \PP(\mathcal{E}_2) (c_1 - \EE[\Delta_1 \mid \mathcal{E}_2]) +  \PP(\mathcal{E}_3) (-c_2 - \EE[\Delta_2 \mid \mathcal{E}_3]). 
\end{align*}
When the right-hand side of the above inequality is larger than zero, the expected total cost of $\Continue$ is larger than the expected cost of $\Stop$. Therefore, we obtain a sufficient condition to $\Stop$ at the current step. 

To continue the analysis, we assume that we have an almost-sure upper bound $\Delta$ on $\EE[\Delta_1 \mid \mathcal{E}_2]$ and $\EE[\Delta_2 \mid \mathcal{E}_3]$: 
\[
    \EE[\Delta_1 \mid \mathcal{E}_2] \leq \Delta \textit{ a.s.} \text{ and } \EE[\Delta_2 \mid \mathcal{E}_3] \leq \Delta \textit{ a.s.}.
\]
A naive bound for $\Delta$ is the upper bound of $C$, e.g., $\max N_{\text{target}}\cdot t_{\text{target}} + \max N_{\text{draft}}\cdot t_{\text{draft}}$. We assume that both the maximum generated tokens and the numbers of candidate tokens per round have an upper limit, so the upper bound is finite. 

Then
\begin{align*}
    & \PP(\mathcal{E}_1) c_1 +  \PP(\mathcal{E}_2) (c_1 - \EE[\Delta_1 \mid \mathcal{E}_2]) +  \PP(\mathcal{E}_3) (-c_2 - \EE[\Delta_2 \mid \mathcal{E}_3]) \geq 0 \\
    \Leftrightarrow & \quad \PP(\mathcal{E}_1) c_1 +  \PP(\mathcal{E}_2) c_1 \geq   \PP(\mathcal{E}_3)c_2  + \PP(\mathcal{E}_3) \EE[\Delta_2 \mid \mathcal{E}_3] + \PP(\mathcal{E}_2)\EE[\Delta_1 \mid \mathcal{E}_2]\\
    \Leftarrow & \quad \PP(\mathcal{E}_1) c_1 +  \PP(\mathcal{E}_2) c_1 \geq   \PP(\mathcal{E}_3)c_2  + \PP(\mathcal{E}_3) \Delta + \PP(\mathcal{E}_2) \Delta\\
    \Leftarrow & \quad \PP(\mathcal{E}_1) c_1  \geq  ( \PP(\mathcal{E}_2) + \PP(\mathcal{E}_3))c_2  + (\PP(\mathcal{E}_3) + \PP(\mathcal{E}_2)) \Delta\\
    \Leftrightarrow & \quad \PP(\mathcal{E}_1) \geq \frac{c_2 + \Delta}{c_1 + c_2 + \Delta}.
\end{align*}
Finally, we note that 
\begin{align*}
\PP(\mathcal{E}_1) &= \PP[\exists 1 \leq i \leq k+1, \text{ such that } Y_i \text{ is rejected} \mid Y_1,\dots,Y_k] \\
&\geq \PP[\exists 1 \leq i \leq k, \text{ such that } Y_i \text{ is rejected} \mid Y_1,\dots,Y_k],
\end{align*}
which concludes the proof.
\end{proof}

\end{document}